\documentclass[]{elsarticle}

\usepackage{graphicx}
\usepackage{subfig}
\usepackage{amssymb}

\usepackage{amsmath}
\usepackage{url}
\usepackage{color}
\usepackage{enumitem}
\usepackage[usenames,dvipsnames,svgnames,table]{xcolor}
\usepackage{multirow}
\usepackage{float}
\definecolor{darkgreen}{rgb}{0.0, 0.5, 0.0}

\usepackage{array}   

\usepackage{algorithmic}
\usepackage{graphicx}
\usepackage{textcomp}\usepackage[dvipsnames]{xcolor}
\usepackage{amssymb}
\usepackage{pifont}
\usepackage{comment}
\usepackage{todonotes}
\usepackage{booktabs}
\usepackage{xltabular}
\usepackage{hyperref}

\usepackage[normalem]{ulem}
    
\journal{Artificial Intelligence in Medicine}

\begin{document}

\begin{frontmatter}


\title{Explanatory Argument Extraction of Correct Answers in Resident Medical Exams}



\author{Iakes Goenaga, Aitziber Atutxa, Koldo Gojenola, Maite Oronoz, Rodrigo Agerri}

\address{HiTZ Center - Ixa, University of the Basque Country UPV/EHU\\  \texttt{\{iakes.goenaga, aitziber.atucha, koldo.gojenola, maite.oronoz,rodrigo.agerri\}@ehu.eus}}

\begin{abstract}
Developing the required technology to assist medical experts in their everyday
activities is currently a hot topic in the Artificial Intelligence
research field. Thus, a number of large language models (LLMs) and
automated benchmarks have recently been proposed with the aim of facilitating
information extraction in Evidence-Based Medicine (EBM) using natural language
as a tool for mediating in human-AI interaction. The most representative
benchmarks are based on a question answering setting, where the objective is
for the models to provide answers to medical questions. Although interesting,
these benchmarks are limited to either multiple-choice or long-form answers
and are available only in English. In order to address these shortcomings, in this
paper we present a new dataset which, unlike previous work: (i) includes not
only explanatory arguments for the correct answer, but also arguments to reason
why the incorrect answers are not correct; (ii) the explanations are written
originally by medical doctors to answer questions from the Spanish Residency
Medical Exams. Furthermore, this new benchmark allows us to setup a novel
extractive task which consists of \emph{identifying the explanation of the
correct answer written by medical doctors}. An additional benefit of our
setting is that we can leverage the extractive QA paradigm to automatically
evaluate performance of LLMs without resorting to costly manual evaluation by
medical experts. Comprehensive experimentation with language models for Spanish
shows that sometimes multilingual models fare better than monolingual ones, even
outperforming models which have been adapted to the medical domain. Furthermore, results across the monolingual models are mixed, with supposedly smaller and inferior models
performing competitively. In any case, the obtained results show that our novel
dataset and approach can be an effective technique to help medical practitioners in identifying relevant evidence-based explanations for medical questions.
 
\end{abstract}


\end{frontmatter}


\section{Introduction}
 
There is an increasing interest in the medical domain to apply Artificial
Intelligence (AI) techniques to assist medical experts in their everyday
activities. These include automatic information extraction methods to obtain,
from large amounts of unstructured text, relevant and timely information to
facilitate the medical experts' deliberation and decision processes as part of
the Evidence-Based Medicine (EBM) practice. In other words, the goal is to help
in EBM which is understood as the ``conscientious, explicit, and judicious use of
current best evidence'' \cite{Sackett1996EvidenceBM} to guide clinical
decision-making with scientific information extracted from systematic reviews
of previous work. 

However, devising algorithms to help in clinical decision-making, in which the
medical expert needs to identify and diagnose a disease and prescribe a
treatment based on the available evidence, is far from trivial. First, doing so
requires for the systems to automatically identify, access and correctly
leverage the relevant medical knowledge. Second, the acquired medical knowledge
needs to be adequately used to help disambiguate between the variety of
symptoms, each of which may be indicative of multiple diseases. Finally, the
system must interact with the medical experts in a natural manner, ideally
using natural language.

Recently, a number of works have been proposed in this line of
research using natural language as a tool for mediating in AI-human
interactions. These efforts have been focused on: (i) argumentation and
explanation-based approaches to detect, classify and assess the quality of
previously given argumentative structures in medical texts
\cite{mayer2021enhancing}. The general idea is that such argumentative
structures constitute the basis of Evidence-Based Medicine; (ii) the
development of Large Language Models (LLMs) adapted, via a variety of
techniques, to the medical domain. These include encoder models such as
SciBERT, BioBERT or PubMedBERT
\cite{beltagy2019scibert,lee2020biobert,pubmedbert} for discriminative
modelling. These models, although they established new state-of-the-art results
in classification tasks, are typically smaller in scale and scope with respect
to decoder and autoregressive models such as GPT-3 or PALM
\cite{brown2020language,singhal2022large} and their biomedical counterparts
BioGPT \cite{Luo2022BioGPTGP}, SciFive \cite{Phan2021SciFiveAT}, and Med-PaLM \cite{singhal2022large}, designed
for generative language modelling; (iii) the creation of automated benchmarks
with the aim of facilitating information access in EBM. The most popular
benchmarks are based on a question answering (QA) scenario, in which the aim is for
the models to classify or generate answers to medical questions. Among others,
it is worth mentioning MultiMedQA, a multi-task, seven multiple-choice dataset
\cite{singhal2022large} and TruthfulQA, a dataset to measure the trustworthiness of
LLMs \cite{lin2021truthfulqa}.

While interesting, previous work presents the following limitations. First, the
large majority of the proposed benchmarks and LLMs have been developed for
English. Second, most of the medical QA datasets consist of Yes/No or long-form
single answers to multiple-choice questions. Third, despite huge progress in
LLMs, learning to generate answers in the medical domain remains notoriously
difficult, as it requires knowledge about medical concepts and the capacity to
retrieve accurate and reliable information from medical literature, guidelines
and other data sources \cite{singhal2022large}. This means that medical QA
benchmarks remain extremely challenging due to the complexity and diversity of
medical language and the vast amount of available medical information
\cite{singhal2022large,lin2021truthfulqa,Phan2021SciFiveAT}. Fourth, evaluating
generative tasks requires expensive manual effort by medical experts.

In order to address these challenges, in this work we present a new dataset which, 
in contrast to previous work, includes not only
explanatory arguments for the correct answer but
also arguments to explain why the remaining possible answers are incorrect.
Additionally, the dataset is entirely written in Spanish by medical doctors.

This new dataset is leveraged in order to formulate a novel extractive task which,
taking into account the explanations written for both correct and incorrect
answers, consists of \emph{identifying the explanation of the correct
answer}. In other words, the models should respond to medical questions
regarding the correct answer in a multiple-choice setting, by providing the
piece of text in the context which explains why a given answer is correct.
Furthermore, our new benchmark profits from harnessing the extractive
QA paradigm to automatically assess the performance of Language Model-based Systems (LLMs),
effectively eliminating the need for costly manual evaluation conducted by
medical experts.

The data source for our new dataset consists of the Resident Medical Intern exams,
or \emph{Médico Interno Residente} (MIR) in Spanish, similar to exams for
medical specialists in other countries such as the United States Medical
Licensing Examination (USMLE). The aim of the MIR exam is to assess medical
students' knowledge by means of a multiple-choice questionnaire contextualized
by a short clinical case. Answering the MIR exam questions requires applying
evidence-based decision making and critical thinking according to the available
evidence present in the case. The process implies forming hypotheses compatible
with the evidence and discarding some of those hypotheses as new evidence is
collected. During this deliberation it is specially relevant to provide
arguments in favour or against each decision, namely, to offer explanations to
support or attack a given hypothesis. 

Thus, the MIR-based dataset introduced in this paper includes the clinical case, the
correct answer, the multiple-choice questions and the annotated explanations result of
the deliberation process outlined above. These explanations have been written
by native Spanish medical doctors and offer a unique resource to set
up a novel dataset and task for EBM consisting of extracting, from the
argumentative structure, the explanation which corresponds to the correct
answer. Summarizing, the main contributions of this work are the following: 

\begin{itemize}
    \item The first dataset for benchmarking LLMs in the medical domain which
		contains argumentative structure to provide explanations for both
		correct and incorrect answers in a multiple-choice questionnaire.
	\item A novel task based on extractive QA to \emph{identify the explanations of
		the correct answer} in \emph{commented} medical exams, thereby helping medical doctors to
		automatically retrieve relevant evidence from existing resources.
    \item The first dataset for medical QA in a language other than English, facilitating future research on
		multilingual approaches.
	\item By formulating this task in an extractive setting, we can objectively
		and automatically evaluate the performance of LLMs without requiring
		any manual annotation by medical experts.
	\item Comprehensive experimentation with LLMs for Spanish demonstrates that
		sometimes multilingual models fare better than monolingual models, even
		outperforming those which have been adapted to the medical domain. We
		show that results across the monolingual models are mixed, with
		supposedly smaller and inferior models performing competitively.
	\item Our reported results suggest that our approach can be an
		effective technique to help medical practitioners in identifying
		relevant evidence-based explanations for medical questions.
    \item Data and code are publicly available to encourage research on this
		research topic and to guarantee reproducibility of results\footnote{\url{https://github.com/ixa-ehu/antidote-casimedicos}}.
\end{itemize}

Finally, we believe that our work can be a stepping stone for other interesting
tasks based on generative modelling and explainable AI in Evicence-Based
Medicine, for example, to predict not only the correct answer
but also to generate an explanation associated to it.

In the rest of the paper, we first discuss the related work, while in Section
\ref{sec:datasets} we describe the dataset creation and annotation. Section
\ref{sec:experimental-setup} introduces the methodology and experimental setup.
Results are reported in Section \ref{sec:results} and discussed and analyzed in
Section \ref{sec:discussion}. To wrap up, section \ref{sec:conclusion} offers some
concluding remarks and future work.

\section{Related work}\label{sec:related}

Due to the unique characteristics of our new dataset and task, there is not
directly comparable work in the literature. However, by casting our new task in
an extractive Question Answering (QA) setting, we can review related work on QA both from an
extractive and abstractive/generative approaches.

In abstractive QA, generative models are not restricted to the input context to create the answer
of the question as they can generate answers word by word employing the entire
vocabulary in an auto-regressive manner \cite{raffel2020exploring}. However,
although the answers obtained by the models following this approach are
apparently good, they are not always factually reliable. Thus, the TruthfulQA
benchmark \cite{lin2021truthfulqa} tested large pre-trained generative models in order to find out whether the answers
given to the questions are truthful. The main conclusion drawn by the authors
was that the models generated many false answers that mimic popular
misconceptions and have the potential to deceive humans. In fact, an
interesting result from that paper was that the larger model, the least
truthful it was. 

In extractive Question Answering the objective is, given a textual context, to
extract a span from the context which constitutes the answer to the given
question \cite{rajpurkar-etal-2016-squad,fisch-etal-2019-mrqa}. This type of QA
is a very popular setting which has seen a high activity in the development of
both datasets specifically about QA
\cite{rajpurkar-etal-2016-squad,rajpurkar-etal-2018-know,reddy2019coqa,
kwiatkowski2019natural}, about reading comprehension
\cite{yang2015wikiqa,lai2017race,zellers2018swag}, and on techniques specific
to this task
\cite{raffel2020exploring,hovy2000question,moreda2011combining,bordes2014question,devlin2018bert,
shao2019transformer}. 

A number of interesting works on QA for the medical domain can be found in the
literature. Many datasets of various characteristics have been developed 
\cite{jin2019pubmedqa,abacha2017overview,vilares2019head,abacha2019bridging,jin2021disease,singhal2022large,
pal2022medmcqa} and workshops and shared tasks such as the
\emph{Question Answering in the medical domain} \cite{abacha2019overview} \&
BioASQ Challenge
\cite{nentidis2020results,nentidis2020overview,nentidis2021overview} have been
organized. As it has been been the case for most NLP tasks,
Transformer-based Pre-trained Language Models (PLMs) have facilitate huge
improvements in the state-of-the-art for medical QA \cite{jin2021disease,ngai2021transformer,yoon2022sequence,pal2022medmcqa}. The
most popular approaches are those that use PLMs pre-trained on medical
corpora such as SciBERT \cite{beltagy2019scibert}, BioBERT
\cite{lee2020biobert} or PubMedBERT \cite{gu2021domain} Furthermore, approaches
harnessing general purpose datasets such as the Stanford Question
Answering Dataset (SQuAD) \cite{rajpurkar-etal-2016-squad} or the Spanish Question Answering
Corpus (SQAC) \cite{gutierrez2021spanish} as intermediate fine-tuning dataset
before training on domain-specific medical datasets have also been shown to be effective
\cite{maximo2022supervised,rosa2022overview}.
 
Nevertheless, despite the extensive efforts of the research community,
question answering for medical exams remains a formidable
challenge, largely due to the scarcity of suitable datasets that must be
annotated by medical specialists. Although there are several datasets in which
doctors' explanations of the correct answer to the question are included such
as, MultiMedQA \cite{singhal2022large}, MedQA \cite{pal2022medmcqa}, PubMedQA \cite{jin2019pubmedqa}, LiveQA
\cite{abacha2017overview} or MedicationQA \cite{abacha2019overview}, most of
them consist of Yes/No or long-form single answers to multiple-choice questions.

However, unlike the MIR-based dataset that we present in this work, none of
them include explanations justifying both the right answer and arguing about
the incorrect ones. Our new dataset not only enables the application of Extractive
QA on explanations given by medical specialists, but it may also open the door to
create clinical argumentative models as they allow to identify the piece of
argument supporting the right answer among arguments explaining why the
remaining answers are incorrect.

\section{Datasets}\label{sec:datasets}

As mentioned in the introduction, one of the most important contributions of
this work is the release of a dataset based on MIR exams commented by medical
doctors. We identified two data sources to build our dataset, namely, the
\emph{CasiMedicos} and the
\emph{MIR Asturias} projects. 
\begin{table}
\centering
\scriptsize
\begin{tabular}{l |p{11.0cm}}
\toprule
\multicolumn{1}{c}{} & \multicolumn{1}{c}{\textbf{Example of a document from the CasiMedicos Dataset}} \\ \midrule 
\textbf{C} & A 30-year-old woman comes to your surgery because she notices a recent lump in her right breast. Her grandmother had breast cancer. On examination, a 2.5 cm nodule with regular borders was palpated in the upper outer quadrant. She has no previous imaging tests. \\ \midrule
\textbf{Q} & Mark the correct answer: \\ \midrule
\textbf{P} & \begin{tabular}[c]{@{}p{11.0cm}}\textbf{(1)} You request a breast ultrasound given the patient's age. \\ \textbf{(2)} You order a mammogram because it is the "gold standard" diagnostic test. \\ \textbf{(3)} You order an MRI given the patient's age and family history.\\ \textbf{(4)} Explains that, as she is 30 years old, it is probably a cyst and will opt for a clinical check-up in 6-12 months. If it persists, then he will order an imaging test.\end{tabular}  \\
\midrule
\textbf{E} & We are presented with a 30 year old female patient presenting with a 2.5 cm nodule with regular borders. The clinical suspicion is of a benign lesion. The imaging test to be performed depends on the age of the patient, the clinical suspicion and the time elapsed since the last examination. \textbf{In patients \textless 35 years of age it is recommended to start the breast study with ultrasound, adding mammography if there are criteria for malignancy.} MRI is reserved for patients with BRCA, breast prostheses, a history of neoadjuvant or conservative surgery. In the case of a palpable lesion of recent appearance and in the absence of any previous imaging test, an ultrasound study is recommended rather than a check-up in 6-12 months. \\ 
\midrule
\textbf{A}  & In patients \textless 35 years of age it is recommended to start the breast study with ultrasound, adding mammography if there are criteria for malignancy.\\ \bottomrule
\end{tabular}
\caption{Example of a document in the CasiMedicos dataset with the correct explanation span manually annotated. \textbf{C}: Clinical Case;
\textbf{Q}: Question; \textbf{E}: Medical Doctor's Explanations; \textbf{P}:
Possible Answers; \textbf{A}: Correct Answer Explanation. The \emph{Clinical Case}, \emph{Question}, and \emph{Possible Answers} sections are generated for the MIR exams by the Spanish Ministry of Health. The medical doctors' explanations (\textbf{E}) and the manually annotated span identifying the explanation of the correct answer (\textbf{A}) are contributions of this work.}
\label{tab:documentexample}
\end{table}

Table \ref{tab:documentexample} shows an example document from our dataset. It
should be noted that the original data is written in Spanish, but for
illustrative purposes we offer a translation into English obtained with
DeepL\footnote{\url{https://www.deepl.com/en/translator}}. This particular
example is taken from the \emph{CasiMedicos} set, although both resources are
structured in the same manner.

The first three parts of the example document, clinical case (\textbf{C}),
question (\textbf{Q}), possible answers (\textbf{P}), plus the solution (out of
4 options), are made public every year by the Spanish Ministry of Health. The
last two parts, namely, the comments or medical doctor's explanations
(\textbf{E}) and the annotation of the span identifying the correct answer
explanation (\textbf{A} and in bold in the \textbf{E} component) are the
contribution of this work. More specifically, the explanatory argumentation
(\textbf{E}) has been written by Spanish medical doctors on top of which we
manually annotated the \emph{span corresponding to the explanation of the
correct answer}.

For each of the questions in both data sources, CasiMedicos and MIR Asturias,
we gathered the year, a unique identifier, whether it had been canceled in the
official exam, the specialty related to the question (digestive, surgery,
pediatrics\ldots), the question itself (\textbf{Q}), the possible answers
(\textbf{P}), the correct answer and the explanations written in the commentary
given by a medical doctor indicating the reasons for choosing or excluding the
answers (\textbf{E}).

The MIR exams contain three types of questions: (i) those associated with an
image (usually a radiography) which needs to be interpreted in order to answer
the question; (ii) questions including a short clinical case as contextual
information previous to the question; (iii) questions about general medical
knowledge without any specific context.

For this work we selected questions of type (ii), namely, those that include
a clinical case (\textbf{C} in the example) to contextualize the question,
possible answers and the explanations given.

The process of manually annotating the corpus consisted of specifying where the
explanation of the correct answer (\textbf{A}) begins and ends (marked in bold
in the \textbf{E} part). In order to obtain grammatically complete correct
answer explanations, annotating full sentences or subordinate clauses was
preferred over shorter spans. The annotation took the equivalent of a person's
month work (4 weeks, 160 hours). In the following we provide a description for each of the data sources used to
build our datasets.


\noindent \textbf{CasiMedicos} is a community of medical professionals who collaboratively,
voluntarily, and free of charge, publishes written explanations about the
possible answers included in the MIR exams. The aim is to generate a resource
that helps future medical doctors to study towards the MIR examinations. The
commented MIR exams, including the explanations, are published in the
CasiMedicos \emph{Project MIR 2.0}
website\footnote{\url{https://www.casimedicos.com/mir-2-0/}}. 

After crawling, cleaning, and organizing the data, we obtained 1,561
commented questions corresponding to the years 2011-2014, 2016, and 2018-2022. 
Selecting those questions which included \emph{clinical cases} reduced the number of
documents from 1,561 to 575. The next step was to create three splits for
training, development and testing (70\%, 15\% and 15\%, respectively).
Table \ref{tab:datasets-statistics} provides some statistics regarding the
final dataset where it can be observed that pediatrics is the most common specialty 
and that clinical case entries are longer (185 words wrt to the overal average
of 163).

\noindent \textbf{MIR Asturias} is a well-known private academy in Spain which
offers courses to prepare for the MIR exams. They produce a number of
interesting material, including textbooks consisting of the commented exams for
the MIR exam of the previous year. We collected 10 books (from 2010 and between
2012-2020) which were publicly available in their website\footnote{\url{https://www.curso-mir.com}}. They contained a total of
2,243 questions with associated explanations written by specialist doctors.
Being a professionally produced resource, the explanations are written following
a clear structural and focused pattern.

After cleaning and pre-processing the data, selecting the questions including
clinical cases left a total of 778 documents.  Table
\ref{tab:datasets-statistics} shows that that the number of words per entry is
slightly higher in MIR Asturias (250 words vs. 185 in \emph{CasiMedicos}) and
that the most repeated specialty is, in this case, Digestive.



\begin{table}[!htbp]
\centering
\begin{tabular}{l |r|r}
\toprule
\textbf{CasiMedicos} & \textbf{Full} & \textbf{ Clinical Cases} \\ \midrule
\#docs & 1,561 & 575 \\ 
\#words & 254,655 & 108,497 \\ 
\#word avg. in docs & 163 & 185 \\ 
\#docs top speciality & Pediatrics (42)  & Pediatrics (41) \\ 
\midrule
\textbf{MIR Asturias}   & \textbf{Full} & \textbf{Clinical Cases} \\ \midrule
\#docs  & 1,971     & 778   \\
\#words  & 450,495 & 199,161 \\
\#word avg. in docs & 228    & 258    \\
\#docs top speciality   & Digestive (157) & Digestive (72) \\ \bottomrule

\end{tabular}
\caption{Statistics of the \emph{CasiMedicos} and \emph{MIR Asturias} datasets. The \emph{Clinical
Cases} subset includes our annotations to identify the spans referring to the
explanation of the correct answer.}
\label{tab:datasets-statistics}
\end{table}

Some issues worth mentioning which differentiate the datasets obtained
from CasiMedicos and MIR Asturias. First, the writing style is slightly more
spontaneous and heteregeneous in CasiMedicos. This is because, unlike MIR
Asturias, CasiMedicos is the result of a voluntary effort by medical doctors in
Spain to provide the comments to the yearly MIR exams. Thus, the length of the
explanations can vary across questions, as well as the structure of the
explanation itself, which does not obey a clear writing pattern or predefined
methodology. In contrast, the comments given for
the MIR Asturias data are much more structured, often with explicit clues as to
which of the possible answers the explanation is referring to. This may also
explain why the average length of the MIR Asturias documents is higher than in
CasiMedicos. Second, the set of questions included in each of the datasets is
different. As a community-based voluntary effort, and unlike MIR Asturias,
CasiMedicos does not include comments for every question in the MIR exams
published every year by the Spanish Ministry of Health. 
Third, contrary to the data obtained from \emph{CasiMedicos} dataset, we did not
manage to obtain permission to share the \emph{MIR Asturias} data, not even for
research purposes. Hence, only CasiMedicos is made publicly available\footnote{We repeatedly contacted MIR Asturias without any success.}.

\section{Materials and Methods}\label{sec:experimental-setup}

Since the apparition of Pre-trained Language Models (PLMs)
\cite{devlin2018bert}, a commonly used approach to solve many NLP tasks has
been to adapt a PLM to the task at hand by fine-tuning it using a task specific
dataset. Pre-trained Language Models encode information about language syntax
and semantics, and the fine-tuning phase provides the model with additional
information specific to the final target task. By adopting this approach, it is
possible to generate models that are customized for a particular task while
simultaneously leveraging the benefits of an already-existing, task-agnostic
Language Model. 

Supplementary Training on Intermediate Labeled-data Tasks (STILT)
\cite{Phang2018SentenceEO} is a variation of the fine-tuning approach, whereby
the pre-trained Language Model (PLM) undergoes an initial fine-tuning on an
intermediate task prior to being fine-tuned on the final downstream target
task. In this case the intermediate task will be extractive QA where the goal
is, given a question, to correctly extract the textual span corresponding to
the right answer. 

\begin{figure}
\includegraphics[width=\textwidth,height=\textheight,keepaspectratio]{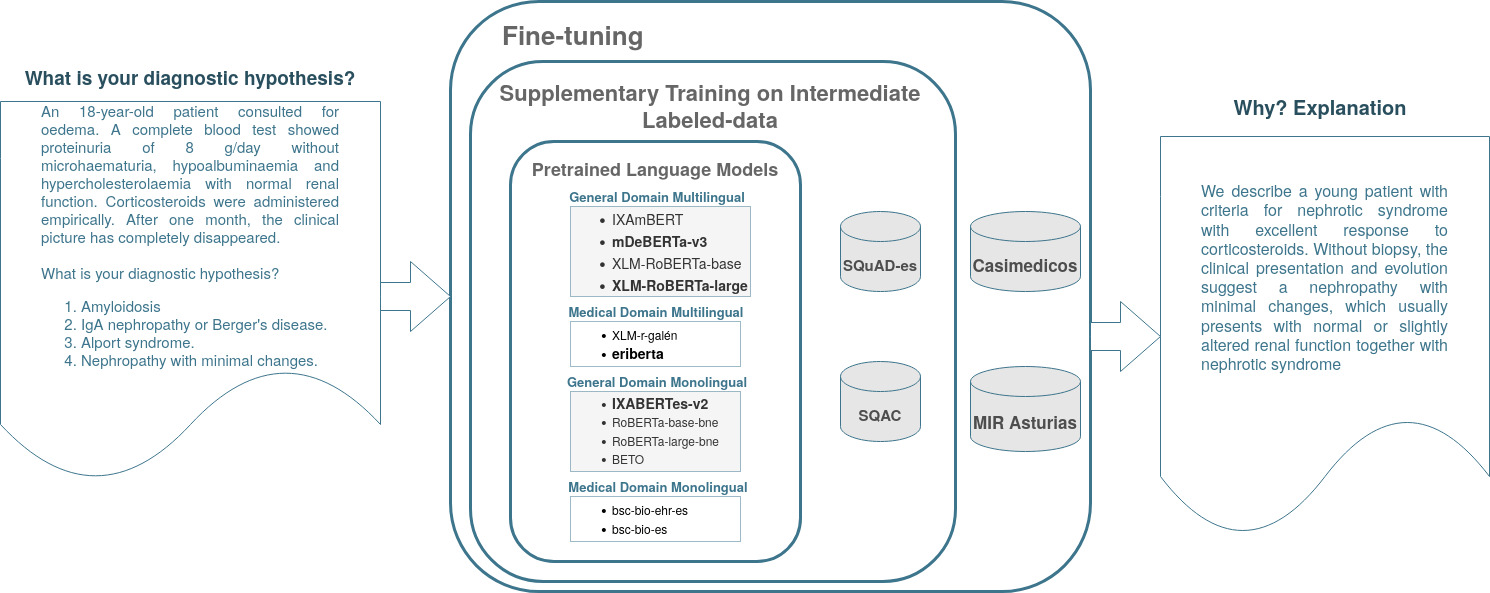}
\caption{General overview of the system}
\label{fig:eskema}
\end{figure}

Figure \ref{fig:eskema} presents a general overview of the system. We first
fine-tune various PLMs on intermediate labeled data, consisting of general QA
Spanish datasets (SQuaD-es and SQAC). This intermediate step may consist of
fine-tuning on either SQUAD-es or SQAC or on their concatenation. The final
step is to then fine-tune the model on the downstream final task, namely, on
either the \emph{CasiMedicos} or the \emph{MIR Asturias} datasets.
In the rest of this section we provide an account of the \emph{fine-tuning
datasets used for STILT} and of the \emph{PLMs} used in the experimentation.

\subsection{Intermediate Fine-tuning Datasets for Extractive QA}\label{Finedata}

The Intermediate Fine-tuning Datasets are based on the English SQuAD extractive
Question Answering setting. While there are other interesting datasets
\cite{reddy2019coqa}, SQuAD has been and remains the main benchmark for Reading
Comprehension, which is perhaps the reason for it being translated into
multiple languages, including Spanish, the target language in this paper.

SQuAD \cite{rajpurkar-etal-2016-squad,rajpurkar-etal-2018-know} is a general domain reading comprehension dataset,
consisting of more than 100,000 questions posed by crowdworkers on a set of
Wikipedia articles, where the answer to every question is either a segment of
text, or span, from the corresponding reading passage, also known as the
\emph{context}. While in SQuADv1 \cite{rajpurkar-etal-2016-squad} every question had an associated textual span
as an answer, SQuADv2 \cite{rajpurkar-etal-2018-know} included an answer for 66\% of
the questions while the remaining 33\% were unanswerable.

Table \ref{tab:squadadib} presents an example of the most relevant parts of a
SQuAD document, namely, the Question, Context and Answer. It can also be
observed that the answer's length is 3 words.  In fact, the extracted answers
in SQuAD are in general very short spans and quite to the point, averaging 3.2
words per answer of which 35.9\%  correspond to named entities, 25\% to
noun phrases and 16.5\% to numerical expressions \cite{reddy2019coqa}.

\begin{table}
\centering
\footnotesize
\begin{tabular}{l|p{9.0cm}}
\toprule
\textbf{Question} & In which R\&B group was she the lead singer?  \\ \midrule
\textbf{Context}  & Beyoncé Giselle Knowles-Carter (born 4 September 1981) is
an American singer, songwriter, producer and actress. Born and raised in
Houston, Texas, she performed in several singing and dancing competitions as a
child, and rose to fame in the late 1990s as the lead singer of the all-female
R\&B group \textbf{Destiny 's Child}. Led by her father, Mathew Knowles, the group
became one of the best-selling girl groups of all time. Her hiatus saw the
release of Beyoncé's debut album, Dangerously in Love (2003), which established
her as a worldwide solo artist, won five Grammy Awards and featured the
Billboard Hot 100 number one single "Crazy in Love". \\ \midrule
\textbf{Answer}   & \textbf{text:} Destiny 's Child, \textbf{answer\_start:} 306 \\
\bottomrule
\end{tabular}
\caption{An example of SQuAD-es dataset document translated to English.}
\label{tab:squadadib}
\end{table}

As shown by Table \ref{tab:documentexample}, in this work we aim to solve the
novel task of extracting, from a context including also explanations about the
wrong answers, the textual span corresponding to the supporting explanatory
argument of the right answer. Thus, although we cast our task in a extractive QA setting, 
our final objective is to extract the explanation for the right answer.
Therefore, in addition to the domain of our datasets
and SQuAD (based on Wikipedia text), perhaps the main difference between
CasiMedicos and MIR Asturias with respect SQuAD is the length of the answers.

In fact, the answer spans in SQuAD are typically much shorter than in CasiMedicos and
MIR Asturias. As it can seen in the SQuAD example given by Table \ref{tab:squadadib},
answers in SQuAD are very short, averaging 3.2 words per answer, compared to
32.02 words average per \emph{explanation of the correct answer} in
CasiMedicos. This is illustratd by the example in Table
\ref{tab:squadadib} where the answer, \emph{Destiny's Child}, a named
entity, is much shorter than most of the explanations to the correct answer
annotated in CasiMedicos (e.g., Table \ref{tab:documentexample} which consists of 
24 words). Thus, while in SQuAD the answers are mostly named entities, short noun phrases
or numerical expressions, in CasiMedicos the explanations span usually correspond to
one or more full sentences (or at least a full subordinate clause), making it a
much more challenging task than extracting very short answers, as in SQuAD.

In this paper we use two extractive Spanish QA datasets as intermediate
fine-tuning datasets in a STILT approach, namely, SQuAD-es and SQAC.

\textbf{The SQuAD-es} dataset \cite{Casimiro2019} is the Spanish
version of SQuAD \cite{rajpurkar-etal-2016-squad,rajpurkar-etal-2018-know}. In
order to create the Spanish version of the dataset, \citet{Casimiro2019}
developed the Translate Align Retrieve (TAR) method to automatically translate
SQuAD into Spanish.
 
\textbf{The Spanish Question Answering Corpus (SQAC)}
\cite{gutierrez2021spanish} is a general domain corpus built for
extractive QA with only answerable questions, as in SQuAD v1. It was created from texts
extracted from the Spanish Wikipedia, encyclopedic articles, newswire articles
from Wikinews, and the Spanish section of the AnCora corpus
\cite{taule2008ancora}. It consists of 18,817 questions with the annotation of
their answer spans from 6,247 textual contexts, following the guidelines
originally developed to built SQuAD v1.1. 

\subsection{Model Building}\label{Modelbuilding}

For the experimentation various PLMs have been selected following certain
criteria such as: domain (medical vs general), size (base vs large) or
multilinguality (multilingual vs monolingual). The choice of models from the
general general domain, both multilingual and monolingual is based on
previously published results on the evaluation of Spanish language models
\cite{Agerri2022LessonsLF}. With respect to the PLMs fine-tuned or pre-trained
for the medical domain, we picked all the models available (up to our
knowledge). The final selection of PLMs consists of the following 12
models:

\begin{itemize} \item \textbf{General Domain Multilingual:} IXAmBERT
	\cite{otegi2020conversational}, mDeBERTa-v3-base \cite{he2021debertav3},
	XLM-RoBERTa (XLM-R) base and large \cite{DBLP:journals/corr/abs-1911-02116}. 
    \item \textbf{Medical Domain Multilingual:} XML-R-Gal{é}n
		\cite{lopez2021transformers}, eriBERTa \cite{delaiglesia2023eriberta}
	\item \textbf{General Domain Monolingual:} IXABERTes-v2
		(IXAes)\footnote{\scriptsize\url{http://www.deeptext.eus/resources/ixabertes-v2.zip}},
		RoBERTa-base-bne (MarIA-base) and RoBERTa-large-bne (MarIA-large)
		\cite{gutierrez2021spanish}, and BETO \cite{CaneteCFP2020}.
    \item \textbf{Medical Domain Monolingual:} bsc-bio-ehr-es and bsc-bio-es \cite{carrino-etal-2022-pretrained}.
\end{itemize}

Table \ref{tab:spanish-lm} provides the main characteristics of the PLMs used
in this paper, including the corpus type and size on which
they were trained, and technical pre-training details such as the number of
layers, the hidden size, number of attention heads, the vocabulary and the
number of parameters.

With respect to the \emph{general domain PLMs}, BETO and IXAmBERT are BERT-base
models pre-trained with both Masked Language Modeling (MLM) and Next Sentence
Prediction (NSP) \cite{devlin2018bert}. BETO was trained for 2M steps in two
different stages: 900K steps with a batch size of 2048 and maximum sequence
length of 128, and the rest of the training with batch size of 256 and maximum
sequence length of 512.  IXAmBERT was trained by executing 1M steps with 256
batch size and 512 sequence length.

IXAes and the MarIA models (base and large) are based on the BERT architecture
but following the RoBERTa procedure \cite{Liu2019RoBERTaAR}: (i) trained only
on the MLM task, (ii) on larger batches (iii) on longer sequences and (iv),
with dynamic mask generation. IXAes performed 120.500 steps with 2048 batch
size and sequence length 512. However, the MarIA models opted instead for a
batch of 2048 and 512 sequence length, but reducing the training to one epoch
without dropout \cite{Komatsuzaki2019OneEI}.

Moving on to the multiligual models, XLM-R base and large were trained
over 1.5M steps with batch 8192 and sequences of 512 length, while mDeBERTa
\cite{he2021debertav3} is also based on RoBERTa but incorporating
disentangled attention, gradient-disentagled embedding sharing and, most
importantly, replacing the MLM task with replaced token detection
\cite{Clark2020ELECTRAPT}; mDeBERTa was trained following the XLM-RoBERTa
procedure but reducing the steps from 1.5M to 500K.

Finally, the \emph{PLMs adapted to the medical domain}: XLM-R-gal\'en was
further pre-trained following the XLM-R method described above on a corpus of
unlabeled oncology clinical texts \cite{lopez2021transformers}. The bsc-bio and bsc-bio-ehr monolingual models were pre-trained following the RoBERTa method, running the training for 48 hours using Adam optimizer with a peak learning rate of 0.0005, 10,000 warm-up steps and a batch size of 2,048 sentences.

Thus, while the specific pre-training details and the corpora used to generate
the PLMs are substantially different the fine-tuning performed to evaluate the
models on the intermediate and final downstream task will follow the same
methodology.

\begin{table*}
\centering
\small
\begin{tabular}{l|llcrrrr}
\toprule
	Model & corpus & \#words & L & H & A & V & \#params \\ \midrule
    IXAmBERT & Wiki & 0.7B & 12 & 768 & 12 & 119K & 110M \\
    mDeBERTa-v3 & CC-100 & 9.3B & 12 & 768 & 12 & 250K & 198M \\
	XLM-R-base & CC-100 & 9.3B & 12 & 768 & 12 & 250K & 270M \\
	XLM-R-large & CC-100 & 9.3B & 24 & 1024 & 16 & 250K & 550M \\
 	XLM-R-Gal{é}n & CC-100 +  &9.3B + & 12 & 768 & 12 & 250K & 270M\\
 	 &  ~EHRSp & ~64.4M &  &  &  &  & \\
    eriBERTa & MedSpEn& 900M & 12 & 768 & 12 & 50K & 125M \\
	IXAes & OSCAR & 25B & 12 & 768 & 12 & 50K & 125M \\
	MarIA-base & BNE & 135B & 12 & 768 & 12 & 50K & 125M \\
	MarIA-large & BNE  & 135B & 24 & 1024 & 16 & 50K & 350M \\
    BETO & Opus,Wiki & 3B & 12 & 768 & 12 & 30K & 110M \\
    bsc-bio-ehr & EHRMedSp & 1.2B & 12 & 768 & 12 & 50K & 125M \\
    bsc-bio & MedSp & 1.1B & 12 & 768 & 12 & 50K & 125M \\
\bottomrule
\end{tabular}
\caption{Language Models used in this paper. L: layer size; H: hidden size; A:
attention heads; V: vocabulary. Wiki: Wikipedia; CC-100: Common Crawl;
OSCAR: Open Super-large Crawled Aggregated coRpus; EHRSp: Spanish Electronic
Health Records (EHR); MedSpEn: Spanish and English Medical Corpus; EHRMedSp:
Spanish EHR and Medical Corpus; MedSp: Spanish Medical Corpus.}
\label{tab:spanish-lm}
\end{table*}

\subsection{Evaluation}

For evaluation we use the standard measures of precision, recall, and
F1-score used for the SQuAD benchmark as defined in formulae (\ref{eqs}), where TPS = \emph{correctly identified
sections}, FPS = \emph{incorrectly identified sections} (marked by system and
not present in the annotated gold standard) and FNS = \emph{false negatives},
namely, present in the gold standard and not detected by system. 

\begin{equation}\label{eqs}
\begin{array}{lcl}
{\scriptstyle Precision} &=& \frac{ TPS}{ TPS~+~FPS}\\
{\scriptstyle Recall} &=& \frac{ TPS}{ TPS~+~FNS} \\
{\scriptstyle F-score} &=& \frac{ 2~*~Precision~*~Recall}{ Precision~*~Recall}
\end{array}
\end{equation}

\vspace{0.15cm}

In a QA extractive setting such as SQuAD, these metrics are computed over the individual words in the prediction against those in the correct answer. Precision corresponds to the ratio of the
number of shared words to the total number of words in the prediction, and
recall is the ratio of the number of shared words to the total number of words
in the correct answer.

\section{Experimental Results}\label{sec:results}

\begin{table}
\centering
\small
\begin{tabular}{c|c|c}
\toprule
\textbf{Variant} & \textbf{SQuAD} & \textbf{CasiMedicos} \\ \midrule
                        & Question            & Clinical Case + Question                                 \\ 
                        & Context             & Doctor's Explanations                                    \\ 
\multirow{-3}{*}{CQ\#E}     & Answer              & Correct Answer Explanation                               \\ \hline \hline

                        & Question            & Clinical Case + Question + Possible Answers              \\ 
                        & Context             & Doctor's Explanations                                   \\ 
\multirow{-3}{*}{CQP\#E}     & Answer              & Correct Answer Explanation                               \\ \hline \hline
 
                        & Question            & Question                                                 \\ 
                        & Context             & Clinical Case + Doctor's Explanations                    \\ 
\multirow{-3}{*}{Q\#CE}     & Answer              & Correct Answer Explanation                               \\ \hline \hline
 
                        & Question            & Question + Possible Answers                                                 \\ 
                        & Context             & Clinical Case + Doctor's Explanations \\ 
\multirow{-3}{*}{QP\#CE}     & Answer              & Correct Answer Explanation \\ \bottomrule
\end{tabular}
\caption{Experimental variants with \emph{CasiMedicos} and \emph{MIR Asturias}. \textbf{C}: Clinical Case;
\textbf{Q}: Question; \textbf{E}: Medical Doctor's Explanations; \textbf{P}:
Possible Answers. \textbf{Fragment before \#}: generated question; \textbf{after \#}: generated context from which to extract the correct explanation.}
\label{tab:datasets-variants}
\end{table}

As explained in Section \ref{Modelbuilding}, we have experimented with twelve
pre-trained Transformer-based language models using the SQuAD and SQAC as
intermediate datasets and CasiMedicos and MIR Asturias as the final downstream
task. In our experimental setup, the first experiment involves directly
fine-tuning the PLMs on CasiMedicos and MIR Asturias. Additionally, three other
approaches were explored which involved using STILT prior to the final
target task: using either SQAC or SQuAD-es as intermediate tasks, or
the combination of both datasets. Details of the experimental settings are
provided in the next section.

\subsection{Experimental Setup}

In the SQuAD extractive Question Answering (QA) setting, the input to fine-tune the
PLMs contains at least the following document elements: the \emph{Question},
the \emph{Context} and \textit{Answer}. In order to cast our \emph{correct
answer explanation extraction} task as an extractive QA task, we simply
experiment by placing the document parts depicted in Table
\ref{tab:documentexample} (\emph{Clinical Case, Question, Possible Answers and
Doctor's Explanations}) as the \emph{Question}, \emph{Context} and
\emph{Answer} fields used in the SQuAD setting. 

\begin{table}
\centering
\scriptsize
\begin{tabular}{l|p{10.0cm}}
\toprule
 & \multicolumn{1}{c}{\textbf{CQ\#E}}      \\ \midrule
 
\textbf{Question} & \textbf{An 18-year-old patient consulted for oedema. A
complete blood test showed proteinuria of 8 g/day without microhaematuria,
hypoalbuminaemia and hypercholesterolaemia with normal renal function.
Corticosteroids were administered empirically. After one month, the clinical
picture has completely disappeared.} \textsc{What is your diagnostic
hypothesis?}                    \\ \midrule

     & \multicolumn{1}{c}{\textbf{CQP\#E}}         \\ \midrule
     
\textbf{Question} & \textbf{An 18-year-old patient consulted for oedema. A complete blood test showed proteinuria of 8 g/day without microhaematuria, hypoalbuminaemia and hypercholesterolaemia with normal renal function. Corticosteroids were administered empirically. After one month, the clinical picture has completely disappeared.} \textsc{What is your diagnostic hypothesis?} \emph{Amyloidosis. @@ IgA nephropathy or Berger's disease. @@ Alport syndrome. @@ Nephropathy with minimal changes.}   \\
\midrule
     
\textbf{Context}  & We describe a young patient with criteria for nephrotic
syndrome with excellent response to corticosteroids. Without biopsy, the
clinical presentation and evolution suggest a nephropathy with minimal changes,
which usually presents with normal or slightly altered renal function together
with nephrotic syndrome, and which in 85-90\% of cases resolves with steroid
treatment. The age is the only data that is a bit of a problem, since although
it is the most common cause of idiopathic nephrotic syndrome in children and
adolescents, it is usually recommended to perform a biopsy prior to treatment
in those over 16 years of age, and in any case there is little discussion. \\
\midrule
\textbf{Answer}   & We describe a young patient with criteria for nephrotic
syndrome with excellent response to corticosteroids. Without biopsy, the
clinical presentation and evolution suggest a nephropathy with minimal changes,
which usually presents with normal or slightly altered renal function together
with nephrotic syndrome            \\ \midrule
\end{tabular}
\caption{\emph{CasiMedicos} variants in which the clinical case is included
within the question. The clinical case is marked in \textbf{bold}, the question
in \textsc{small capitals}, and the possible answers are in \emph{italics}.}
\label{tab:dataset-variants-example}
\end{table}

In this sense, we try 4 different combinations to use the clinical case
(\textbf{C}), question (\textbf{Q}), possible answers (\textbf{P}) and the
explanations (\textbf{E}) as Question and Context (the correct answer
explanation \textbf{A} is always the answer) in a SQuAD sense. These variants are described in
Table \ref{tab:datasets-variants}. Thus, in the first variant the question in
the SQuAD scenario consists of the clinical case (\textbf{C}) and the question
(\textbf{Q}). In the second one, we add the possible answers (\textbf{P}),
while for the 4th we remove the clinical case (\textbf{C}); finally, in the third
variant we leave just the question (\textbf{Q}). Regarding the context from
which to extract the explanation of the correct answers, in the first two
variants it includes only the full explanations given in the comment written by
the medical doctors  (\textbf{E}) while for the last two we also add the
clinical case (\textbf{C}).

For illustration purposes, Table \ref{tab:dataset-variants-example} provides an
example of the variants CQ\#E and CQP\#E in which the clinical case (\textbf{C}) is
added together with the question (\textbf{Q}) for CQ\#E and also with the possible answers
(\textbf{P}) in CQP\#E. In both cases the context consists of the doctor's explanations
(\textbf{E}) which includes the \emph{correct answer explanation}
(\textbf{A}) to be extracted.

Putting together the 12 PLMs, 4 datasets (SQuAD-es, SQAC, CasiMedicos and MIR
Asturias) and the 4 experimental variants just presented, we fine-tune the
models using the following procedure. We truncate the maximum total input
sequence length to 384 and the maximum length of an answer that could be
generated to 512. We fine-tuned the models for 20 epochs, set the random seed
for initalization to 42, and kept the default QA values for the other
hyper-parameters \cite{devlin2018bert}. Adding up all the hours needed to train
all the models using a NVIDIA GPU V100 32GB, we get approximately 77 hours,
leaving a carbon footprint of 34.81 (lbs CO2e).  The evaluation results are
reported in the next section.

\subsection{Results}

As we use the SQuAD-es and SQAC data to perform
STILT for each of the models, the final number of
experimental results was too large ($\sim$400) to present it here. Thus, in this section we
report the results from the best two multilingual (base and large versions) and
monolingual models\footnote{There is not a large version of the models adapted
to the medical domain for Spanish.} (from the general and the medical domain,
respectively). While Tables \ref{tab:casimedicos-results} and
\ref{tab:mir-results} provide the results obtained on the \emph{CasiMedicos} and \textit{MIR
Asturias} datasets, respectively, all the results using all twelve models can be
found in \ref{sec:A}.

\begin{table}[H]
\centering
\resizebox{12cm}{!} {
\begin{tabular}{c|l|c|c|c|c|c}
\toprule
 \multicolumn{2}{l}{} & \textbf{CM} & \textbf{SQAC+CM} & \textbf{SQES+CM} & \textbf{ALL} & \textbf{Avg.} \\ \midrule
\multirow{4}{*}{\textbf{IXAes}} & \textbf{CQ\#E}  & 68.18       & 67.15            & 65.90                & 69.74                    & 67.74       \\ 
 & \textbf{CQP\#E} & 66.95       & 67.74            & 68.50                & 64.96                    & 67.04       \\ 
 & \textbf{Q\#CE}      & 62.34       & 67.93            & 64.53               & 60.43                    & 63.80       \\ 
 & \textbf{QP\#CE} & 60.81       & 64.93            & 60.35               & 62.85 & 62.23        \\ \midrule
\multirow{4}{*}{\textbf{eriBERTa} } & 
\textbf{CQ\#E} & 67.56       & 70.01            & 70.49               & 67.58                    & 68.91         \\ 
 & \textbf{CQP\#E} & 67.61       & 65.55            & 67.90                & 68.61                    & 67.41       \\ 
&\textbf{Q\#CE}   & 65.53       & 70.46            & 65.58               & 71.39                    & 68.24         \\ 
& \textbf{QP\#CE} 			  & 67.34       & 64.43            & 64.51 & 70.14 & 66.60        \\ \midrule

\multirow{4}{*}{\textbf{mDeBERTa-v3}} &
\textbf{CQ\#E}   & 66.76       & 69.65            & 68.92               & 66.67                    & 68.00            \\ 
&\textbf{CQP\#E} & 58.16       & 68.32            & 66.33               & 66.10                     & 64.72       \\ 
&\textbf{Q\#CE}   & 65.94       & 66.45            & 68.14               & 69.91                    & 67.61         \\ 
&\textbf{QP\#CE} & 57.55       & 65.15            & 65.50                & 70.18 & 64.59        \\ \midrule

\multirow{4}{*}{\textbf{XLM-R-large}} &
\textbf{CQ\#E}   & \underline{71.74}       & \textbf{74.47} & 72.97               & \underline{71.84} & \textbf{72.75}        \\
& \textbf{CQP\#E} & 69.49    & 71.95            & \underline{73.08}               & 71.76                    & 71.57         \\
& \textbf{Q\#CE}   & 62.39   & 66.55            & 67.86               & 70.18                    & 66.74        \\ 
& \textbf{QP\#CE}  & 57.49   & 69.48            & 73.80                & 70.09 & 67.71        \\ \bottomrule
\end{tabular}
}
\caption{F1-scores (partial match) on the \emph{CasiMedicos} (CM) dataset variants
defined in Table \ref{tab:datasets-variants}. \textbf{C}: Clinical Case;
\textbf{Q}: Question; \textbf{E}: Medical Doctor's Explanations; \textbf{P}:
Possible Answers. \textbf{Fragment before \#}: generated question; \textbf{after \#}: generated
context from which to extract the correct explanation. \textbf{Scores in bold:} Best
overall result; \underline{scores underlined}: best result per dataset used for
fine-tuning. SQES: SQUAD-es.} 
\label{tab:casimedicos-results}
\end{table}

\begin{table}[H]
\centering
\resizebox{12cm}{!} {
\begin{tabular}{c|l|c|c|c|c|c}
\toprule
\multicolumn{2}{l}{} & \textbf{MA} & \textbf{SQAC+MA} & \textbf{SQES+MA} & \textbf{ALL} & \textbf{Avg.} \\ \midrule
\multirow{4}{*}{\textbf{IXAes}} & \textbf{CQ\#E}      & 82.37       & 86.56            & \underline{86.34}               & 84.69                    & 84.99         \\ 
& \textbf{CQP\#E} & 79.98       & \underline{86.85}            & 84.39               & 84.80                     & 84.00       \\ 
& \textbf{Q\#CE}      & 81.69       & 83.34            & 81.58               & 79.11                    & 81.43         \\ 
& \textbf{QP\#CE} & 79.14       & 78.02            & 81.15               & 78.95 & 79.31        \\ \midrule
\multirow{4}{*}{\textbf{eriBERTa}} & \textbf{CQ\#E}  & \underline{84.92}       & 85.89            & 85.25               & 84.50                     & \textbf{85.14}         \\ 
& \textbf{CQP\#E} & 82.66       & 85.45            & 84.33               & 85.42                    & 84.46        \\ 
& \textbf{Q\#CE}      & 82.08       & 83.70             & 84.14               & \textbf{87.03}                    & 84.23       \\ 
& \textbf{QP\#CE} & 80.27       & 82.56            & 84.29               & 83.08 & 82.55         \\ \midrule
\multirow{4}{*}{\textbf{mDeBERTa-v3}}  &
\textbf{CQ\#E}      & 79.36       & 82.57            & 79.40                & 82.23                    & 80.89         \\ 
&\textbf{CQP\#E} & 77.82       & 81.51            & 77.32               & 80.80                     & 79.36       \\ 
&\textbf{Q\#CE}      & 78.26       & 79.51            & 79.99               & 79.87                    & 79.40       \\ 
&\textbf{QP\#CE} & 71.49       & 72.19            & 74.39               & 73.72 & 72.94       \\ \midrule
\multirow{4}{*}{\textbf{XLM-R-large}} & 
\textbf{CQ\#E}      & 82.20        & 84.30             & 85.92               & 82.27                    & 83.67       \\ 
& \textbf{CQP\#E} & 81.25       & 83.63            & 79.68               & 82.78                    & 81.83       \\ 
& \textbf{Q\#CE}      & 81.03       & 80.66            & 82.89               & 80.07                    & 81.16       \\ 
& \textbf{QP\#CE} & 74.49       & 77.15            & 77.46               & 79.36
				& 77.11        \\ \bottomrule
\end{tabular}
}
\caption{F1-scores (partial match) on the \emph{MIR Asturias} (MA) dataset variants
defined in Table \ref{tab:datasets-variants}. \textbf{C}: Clinical Case;
\textbf{Q}: Question; \textbf{E}: Medical Doctor's Explanations; \textbf{P}:
Possible Answers. \textbf{Fragment before \#}: generated question; \textbf{after \#}: generated
context from which to extract the correct explanation. \textbf{Scores in bold:} Best
overall result; \underline{scores underlined}: best result per dataset used for
fine-tuning. SQES: SQUAD-es.}
\label{tab:mir-results}
\end{table}

By looking at the results on \emph{CasiMedicos}, reported in Table
\ref{tab:casimedicos-results}, the first interesting piece of information is that the best
overall performance is obtained by a multilingual model not adapted to the medical domain, namely,
XLM-RoBERTa-large (almost 4 points F1-score averaged across the 4 fine-tuning
settings over the second best model, eriBERTa, specifically developed for the
medical domain). Second, while eriBERTa is slightly better than IXAes
and mDeBERTa-v3-base, differences are relatively small. Third, the best setting for this task in \emph{CasiMedicos} is
CQ\#E, in which the question includes both the clinical case and the
question (the context corresponds to the explanations written by the medical
doctors). In general, including the clinical case as part of the question seems to be beneficial. Fourth, there is not a pattern in the effect of the different intermediate fine-tunings (SQAC, SQuAD-es or SQAC+SQuAD-es); while using
STILT by means of general domain QA datasets always helps for the
downstream task, results are not conclusive as to which of the datasets
provides the largest gains.

The results obtained with \emph{MIR Asturias} leave a different picture.
Thus, by looking at Table \ref{tab:mir-results}, we can see that the model
obtaining the best average score is eriBERTa, although perhaps the most
surprising result is the strong performance of IXAes, a relatively small
general domain model for Spanish. Both eriBERTa and IXAes are slightly
better than XLM-RoBERTa-large ($\sim$1.4 points in F-score) while clearly
outperforming mDeBERTa-v3-base ($\sim$4 points in F-score).

As for \emph{CasiMedicos}, the best dataset variant on average remains CQ\#E
and it also helps including the clinical case in the question. Finally, as it
was also the case in \emph{CasiMedicos}, while a previous fine-tune with SQAC,
SQuAD-es or their combination is beneficial, there is not a systematic winner
as to which QA dataset is the best for the majority of the cases. If we were to
choose one, we would probably choose SQAC, as it produces the largest gains
with the best dataset variant (CQ\#E) for most of the models.

To conclude, it is worth noting that the results are substantially better for
\emph{MIR Asturias}. Several possible explanations are discussed in the next section.

\section{Discussion}\label{sec:discussion}

In this section we discuss and analyze the main results reported in the
previous section. More specifically, we will first analyze the possible reasons
to explain the large difference in performance when we compare the results on
the \emph{CasiMedicos} and \emph{MIR Asturias} datasets.  Furthermore, in
Section \ref{analysis2} we will discuss the most difficult cases we encounted
in the datasets.

\subsection{CasiMedicos vs MIR Asturias}\label{sec:mir-vs-casi}

As illustrated by the results reported in Section \ref{sec:results}, the models
we fine-tuned and evaluated in the same setting obtain substantially better
results for \emph{MIR Asturias}. For example, if we average the results of the 4 best
systems discussed, for \emph{MIR Asturias} the models obtain an average F-score of
81.40 while for \emph{CasiMedicos} the score corresponds to 67.23, namely, 14.17
points lower. After a manual inspection of the \emph{CasiMedicos} and \emph{MIR Asturias}
training data, we formulate two hypotheses to explain this phenomenon: (1) the
length of the explanations of the correct answer, our task objective, is longer
in the \emph{CasiMedicos} dataset, making the task more difficult to learn; (2) the
explanations in the \emph{MIR Asturias} dataset are generated following a much more
structured method, which makes it easier to identify the explanation of the
correct answer.

In order to test the first hypothesis, we calculated the average length of the
explanations of the correct answers for both datasets. Table
\ref{tab:sentence-length-stats} shows that the correct answers' explanations are
in fact shorter in \emph{CasiMedicos}, which means that this hypothesis is not valid.

\begin{table}[H]
\centering
\small
\begin{tabular}{c|r|r}
\toprule
\textbf{Explanation length} & \textbf{CasiMedicos} & \textbf{MIR Asturias} \\
\midrule
\textbf{1} & 51.13\% & 35.89\% \\ 
\textbf{2} & 38.63\% & 31.62\% \\ 
\textbf{3} & 4.54\%  & 13.67\% \\
\textbf{more than 3} & 5.68\% & 18.80\% \\ \bottomrule
\end{tabular}
\caption{Length of explanations of correct answers, in number of sentences, for the \emph{CasiMedicos} and \emph{MIR Asturias} datasets.}
\label{tab:sentence-length-stats}
\end{table}

With respect to our second hypothesis, that the explanations in the \emph{MIR
Asturias} dataset might be written in a more structured manner, the analysis of
the two datasets has led us to conclude that  this is indeed the case.
If we examine Table \ref{tab:casimedicos-mirasturias-comparison}, which provides
different explanations to the same question, one given in the \emph{MIR
Asturias} and the other one in the \emph{CasiMedicos} dataset, it is possible to
observe that \emph{MIR Asturias} follows a rather clear structured pattern where
the spans of the explanations are systematically marked and linked to their corresponding
possible answers (expressions marked in bold in the explanations in Table
\ref{tab:casimedicos-mirasturias-comparison}).

\begin{table}[H]
\centering
\scriptsize
\begin{tabular}{l|p{11.0cm}}
\toprule
\multicolumn{2}{l}{\textbf{ CasiMedicos}}  \\ \midrule
\textbf{E}  & We describe a young patient with criteria for nephrotic syndrome
with excellent response to corticosteroids. Without biopsy, the clinical
presentation and evolution suggest a nephropathy with minimal changes, which
usually presents with normal or slightly altered renal function together with
nephrotic syndrome, and which in 85-90\% of cases resolves with steroid
treatment. The age is the only data that is a bit of a problem, since although
it is the most common cause of idiopathic nephrotic syndrome in children and
adolescents, it is usually recommended to perform a biopsy prior to treatment
in those over 16 years of age, and in any case there is little discussion. \\
\midrule
\textbf{A}   & We describe a young patient with criteria for nephrotic syndrome
with excellent response to corticosteroids. Without biopsy, the clinical
presentation and evolution suggest a nephropathy with minimal changes, which
usually presents with normal or slightly altered renal function together with
nephrotic syndrome.           \\ \midrule
\multicolumn{2}{l}{\textbf{MIR Asturias}} \\ \midrule
\textbf{E}   & Patient with a nephrotic syndrome that responds to corticosteroids, characteristic of minimal change disease \textbf{(answer 4 correct)}. The only data that can make us doubt is the age of the patient since minimal change disease is characteristic of children from 2 to 6 years old. However, it occurs in other age groups including the elderly.  Although amyloidosis produces nephrotic syndrome, it does not respond to corticosteroids \textbf{(incorrect answer 1)}, and there is no data in the clinical case that makes us suspect primary amyloidosis (plasma cell dyscrasia) or secondary amyloidosis (uncontrolled chronic inflammatory disease). IgA nephropathy or Alport syndrome are not characterized by producing nephrotic syndrome or responding to treatment with corticosteroids \textbf{(answer 2 and 3 incorrect)}.
\\ \midrule
\textbf{A} & Patient with a nephrotic syndrome that responds to
corticosteroids, characteristic of minimal change disease \textbf{(answer 4
correct)}.\\ \bottomrule
\end{tabular}
\caption{Comparing the explanations given for the same question (Table \ref{tab:dataset-variants-example}) in both the \emph{CasiMedicos} and MIR Asturias datasets.}
\label{tab:casimedicos-mirasturias-comparison}
\end{table}

In order to find out whether these structural patterns appearing in  \emph{MIR
Asturias} are actually influencing the results obtained by the models for this
dataset, we performed the following experiment. First, we manually removed every expression such as ``(\textbf{answer 4 correct})'' and so on appearing in the \emph{MIR Asturias} dataset. Second, we fine-tuned and evaluated the models with this new \emph{MIR Asturias} version (\emph{unstructured}) and compared the results with those
reported in Table \ref{tab:mir-results}. Thus, in Table \ref{tab:egituragabe}
we can see the averaged results of the models for both versions of \emph{MIR Asturias}
and for \emph{CasiMedicos}. It is quite clear that by removing the patterns from the
explanations in the \emph{MIR Asturias} data the results obtained are similar to those
of \emph{CasiMedicos}. This suggests that models were somehow being helped by such
patterns in order to learn how to extract the explanation of the correct
answer.

\begin{table}[H]
\centering
\small
\resizebox{12cm}{!} {
\begin{tabular}{l|c|c|c}
\toprule
  & \textbf{MIR Asturias} & \textbf{MIR Asturias (u)} & \textbf{CasiMedicos} \\ \midrule
\textbf{IXAes} & 84.99 & 66.56 & 67.74 \\ 
\textbf{eriBERTa} & \textbf{85.14} & \textbf{70.97} & 68.91 \\ 
\textbf{mDeBERTa-v3} & 80.89 & 62.84 & 68.00 \\ 
\textbf{XLM-R-large} & 83.67 & 67.10 & \textbf{72.75} \\ \bottomrule
\end{tabular}
}
\caption{Comparison in terms of averaged results (across MA, SQAC+MA,
	SQuADES+MA, SQAC+SQuADES+MA) for the CQ\#E variant as described in Table
\ref{tab:casimedicos-results}. MIR Asturias (u): unstructured.}
\label{tab:egituragabe}
\end{table}


Motivated by these results, we performed an error analysis and found out
that those language models fine-tuned on the \emph{MIR Asturias - Structured}
data suffer to extract the correct explanations in the following
two cases:

\begin{enumerate}
	\item When pointers to explicit patterns are absent in the context
from which to extract the correct explanation. After all, not every single
document in the \emph{MIR Asturias} dataset is structured in this manner. For these
cases the model does not provide any answer, it just returns either the entire context
or some random text between brackets.
\item When more than one explanation is marked as correct because the
question is about which of the possible answers is FALSE (as illustrated by
Table \ref{tab:mir-false-question-example}), the models often return the
fragment of the explanations in which the word \textit{correct} appears.
\end{enumerate}

\begin{table}[H]
\centering
\footnotesize
\begin{tabular}{l|p{11.0cm}}
  \toprule
\textbf{Q} & A 45-year-old man, previously
healthy, develops a fever and acute gastroenteritis with severe liquid
diarrhoea. Blood tests show sodium 140 mmol/L, potassium 3.2 mmol/L, chlorine
85 mmol/L and bicarbonate 38 mmol/L. Arterial pH is 7.60 and arterial pCO2 42
mmHg. Arterial pH is 7.60 and arterial pCO2 42 mmHg. \emph{Which of the following
statements is FALSE?}
\\ \midrule
\textbf{E}  & Given the elevated pH, with elevated bicarbonate, and
normal pCO2, the patient has metabolic alkalosis \textbf{(answer 1 correct)}.
The anion GAP (140 - 85 - 38) is 17 and therefore elevated \textbf{(answer 3
correct)}, and the picture is accompanied by low chloride and potassium due to
digestive losses from diarrhoea \textbf{(answer 4 correct)}. The difficulty
with this question is that most patients with diarrhoea have metabolic acidosis
and hypokalaemia, yet this patient has metabolic alkalosis. This is because
some forms of diarrhoea, especially ionic diarrhoea, produce isotonic losses
with chlorine content (chlorine-losing diarrhoea) and this induces metabolic
alkalosis through chlorine loss, and through volume loss with secondary
aldosterone activation. The loss of chlorine explains why the GAP anion is
increased. Apart from this detail, the patient has a normal pCO2 (between 35
and 45) and therefore does not have any form of compensation, let alone an
acidosis in which pCO2 should be elevated \textbf{(answer 2 incorrect)}. \\ \midrule
\textbf{A} & The loss of chlorine explains why the GAP anion is
increased. Apart from this detail, the patient has a normal pCO2 (between 35
and 45) and therefore does not have any form of compensation, let alone an
acidosis in which pCO2 should be elevated \textbf{(answer 2 incorrect)}.\\ \midrule
\end{tabular}
\caption{An example of \emph{MIR Asturias} dataset where the task is to extract the explanation of the incorrect answer.}
\label{tab:mir-false-question-example}
\end{table}

These results led us to conclude that learning extractive QA in our setting
using the \emph{unstructured} version of the datasets is the most interesting
strategy to learn models able to generalize for the \emph{explanation extraction}
task. In other words, the explicit pointers about where the explanation of
the correct answer is located provide spurious clues to the language models,
with the additional undesired effect of artificially inflating the results.

\subsection{Impact of Intermediate Fine-tuning with QA Datasets from the General Domain}\label{analysis2}

In a final exercise of error analysis, we wanted to provide a qualitative
explanation of the kind of improvements derived from performing an intermediate
fine-tuning using the general domain QA datasets, namely, SQAC or SQuAD-es. By
analyzing the predictions of the models trained with or without SQAC
before the final fine-tuning step on \emph{MIR Asturias}, we observed the following
points. If fine-tuned with SQAC (summarized in Table \ref{tab:sqacvsma}): 

\begin{enumerate}
	\item IXAes: fewer \emph{empty} answers and more \emph{exact} matches.  
	\item eriBERTa: fewer \emph{incomplete}\footnote{We consider an answer incomplete when it is too short. For instance: \textit{correct)}, \textit{the}, or \textit{some}.} answers and more \emph{exact} matches.
	\item mDeBERTa-v3-base: fewer \emph{empty} and \emph{incomplete} answers.
	\item XLM-R-large: fewer \emph{empty} answers and more \emph{exact} matches.
\end{enumerate}

\begin{table}[H]
\centering
\scriptsize
\begin{tabular}{l|c|c|c} \toprule
    & \textbf{Less Empty} & \textbf{Less Incomplete} & \textbf{More Exact} \\ \midrule
\textbf{IXAes} & \textcolor{darkgreen}{\checkmark}  &  & \textcolor{darkgreen}{\checkmark} \\
\textbf{eriBERTa} &  & \textcolor{darkgreen}{\checkmark} & \textcolor{darkgreen}{\checkmark} \\ 
\textbf{mDeBERTa-v3} & \textcolor{darkgreen}{\checkmark} & \textcolor{darkgreen}{\checkmark} & \\
\textbf{XLM-R-large} & \textcolor{darkgreen}{\checkmark}  & & \textcolor{darkgreen}{\checkmark} \\ \bottomrule
\end{tabular}
\caption{Improvements on the MIR Asturias data when SQAC is used for intermediate fine-tuning for the CQ\#E variant.}
\label{tab:sqacvsma}
\end{table}

Summarizing, in order to generalize better in the \emph{correct answer's
explanation extraction} task it is better to avoid explicit clues that the
models focus on. In this sense, this turns out to be a good result for
\emph{CasiMedicos}, as it is written in a more spontaneous and unstructured manner
than MIR Asturias. Furthermore, results show that using SQAC or SQuAD-es in a
STILT setting helps to systematically improve the results, in particular by
making the models better in avoiding empty and incomplete answers and boosting
the number of exact matches.

\section{Concluding Remarks}\label{sec:conclusion}

In this paper we present a novel dataset based on Spanish MIR exams which
exhibit characteristics not available in
any previous work. More specifically, it includes explanatory arguments for the correct
answer and also argumentative structure to explaning why the remaining answers are incorrect.
Furthermore, while most of the QA-related datasets in the medical domain are
written in English, in our case it has been written in Spanish by medical doctors.

The MIR-based datasets (\emph{CasiMedicos} and \emph{MIR Asturias}) introduced in this paper include the clinical case, the
correct answer, the multiple-choice questions and the annotated explanations result of
an evidence-based deliveration process. We publicly release the annotated
\emph{CasiMedicos} set to encourage research on this topic and facilitate
reproducibility of
results\footnote{https://github.com/ixa-ehu/antidote-casimedicos}.

The MIR-based datasets allow us to propose a novel extractive task which,
taking into account the explanations written for both correct and incorrect
answers, consists of \emph{identifying the explanation of the correct
answer}. In other words, the models should respond to medical questions
regarding the correct answer in a multiple-choice setting, by providing the
piece of text in the context which explains why a given answer is correct.
Experimentally our new benchmark casts this novel task within the extractive
QA, SQuAD-style, paradigm. By doing so we are able to automatically evaluate
the performance of Pre-trained Language Models while avoiding any costly manual
evaluation that may require medical experts. 

Experimental results demonstrate that, compared to SQuAD, our task is more
complex, especially due to the long length of the spans of text to be extracted
(explanations for the correct answer in \emph{CasiMedicos} averaging 32.02 words length vs 3.2 words
per answer in SQuAD). A comprehensive evaluation of Spanish language models shows that often multilingual non-specialized models outperform
monolingual ones, even those which have been pre-trained with large amounts of
medical texts. Still, reported results show that our novel dataset and approach can be an
effective technique to help medical practitioners in identifying relevant
evidence-based explanations for medical questions.

Finally, the unique characteristics of the \emph{CasiMedicos} dataset released with this work may facilitate
a number of research lines involving discriminative and generative work on
argumentation, explanability and truthfulness, to name but a few. Furthermore, by
harnessing recent advances in label projection for sequence labelling tasks
\cite{garcia-ferrero-etal-2022-model,Yeginbergenova2023CrosslingualAM} 
future work may be addressed both from a monolingual and multilingual point of view.

\section*{Acknowledgements}

We thank the CasiMedicos Proyecto MIR 2.0 for their permission to share their
data for research purposes. This work has been partially supported by the HiTZ
center and the Basque Government (Research group funding IT-1805-22). We also acknowledge the funding from the following \\MCIN/AEI/10.13039/501100011033 projects: (i) Antidote (PCI2020-120717-2), and by European Union
NextGenerationEU/PRTR; (ii) DeepKnowledge (PID2021-127777OB-C21) and ERDF A way
of making Europe; (iii) Disargue (TED2021-130810B-C21) and European Union
NextGenerationEU/PRTR. Rodrigo Agerri currently holds the RYC-2017-23647
fellowship \\(MCIN/AEI/10.13039/501100011033 and by ESF Investing in your
future).

\bibliographystyle{model1-num-names}
\bibliography{biblio.bib}

\newpage

\appendix

\section{}\label{sec:A}

In this appendix the results with every pre-trained language model (PLM) are presented.

\begin{table}[h]
\centering
\footnotesize
\resizebox{12cm}{!} {
\begin{tabular}{c|l|c|c|c|c|c}
\toprule
 \multicolumn{2}{l}{} & \textbf{CM} & \textbf{SQAC+CM} & \textbf{SQES+CM} & \textbf{ALL} & \textbf{Avg.} \\ \midrule
\multirow{4}{*}{\textbf{mDeBERTa-v3}}    & 
\textbf{CQ\#E}      & 66.76       & 69.65            & 68.92               & 66.67                    & 68.00            \\ 
& \textbf{CQP\#E} & 58.16       & 68.32            & 66.33               & 66.10                     & 64.72       \\ 
& \textbf{Q\#CE}      & 65.94       & 66.45            & 68.14               & 69.91                    & 67.61         \\ 
& \textbf{QP\#CE} & 57.55       & 65.15            & 65.50                & 70.18 & 64.59        \\ \midrule
\multirow{4}{*}{\textbf{XLM-R-base}}                 & \textbf{CQ\#E}      & 63.94       & 71.62            & 65.62               & 67.91                    & 67.27       \\ 
&\textbf{CQP\#E} & 64.25       & 72.99            & 71.59               & 69.24                    & 69.51       \\ 
&\textbf{Q\#CE}      & 62.39       & 66.55            & 67.86               & 70.18                    & 66.74        \\ 
&\textbf{QP\#CE} & 56.55       & 68.23            & 65.04               & 64.22 & 63.51         \\ \midrule
\multirow{4}{*}{\textbf{XLM-R-large}}               & 
\textbf{CQ\#E}      & 71.74       & 74.47            & 72.97               & 71.84                    & 72.75        \\ 
& \textbf{CQP\#E} & 69.49       & 71.95            & 73.08               & 71.76                    & 71.57         \\ 
& \textbf{Q\#CE}      & 62.39       & 66.55            & 67.86               & 70.18                    & 66.74        \\ 
& \textbf{QP\#CE} & 57.49       & 69.48            & 73.80                & 70.09 & 67.71        \\ \midrule
\multirow{4}{*}{\textbf{XLM-R-GALEN}}                        & \textbf{CQ\#E}      & 64.47       & 70.27            & 67.08               & 70.65                    & 68.11       \\ 
& \textbf{CQP\#E} & 60.37       & 63.49            & 64.79               & 70.22                    & 64.71       \\ 
& \textbf{Q\#CE}      & 56.65       & 52.65            & 59.87               & 60.98                    & 57.53       \\ 
& \textbf{QP\#CE} & 53.67       & 54.97            & 60.74               & 58.12 & 56.87        \\ \midrule

\multirow{4}{*}{\textbf{eriBERTa}} & 
\textbf{CQ\#E}      & 67.56       & 70.01            & 70.49               & 67.58                    & 68.91         \\ 
& \textbf{CQP\#E} & 67.61       & 65.55            & 67.90                & 68.61                    & 67.41       \\ 
& \textbf{Q\#CE}      & 65.53       & 70.46            & 65.58               & 71.39                    & 68.24         \\ 
& \textbf{QP\#CE} & 67.34       & 64.43            & 64.51               & 70.14 & 66.60    \\ \bottomrule
\end{tabular}
}
\caption{F1-score results of the multilingual models on different \emph{CasiMedicos} dataset variants where \textbf{CM} = Fine-tuning on \emph{CasiMedicos} dataset, \textbf{SQAC} = Fine-tuning on SQAC dataset, \textbf{SQES} = Fine-tuning on SQuAD-es dataset. \textbf{C}: Clinical Case; \textbf{Q}: Question; \textbf{E}: Medical Doctor's Explanations; \textbf{P}:
Possible Answers. \textbf{Fragment before \#}: generated question;
\textbf{after \#}: generated context from which to extract the correct
explanation. SQES: SQUAD-es.}
\label{table:casimedicosmulti}
\end{table}

\begin{table}[h]
\centering
\footnotesize
\resizebox{12cm}{!} {
\begin{tabular}{c|l|c|c|c|c|c}
\toprule
 \multicolumn{2}{l}{} & \textbf{CM} & \textbf{SQAC+CM} & \textbf{SQES+CM} & \textbf{ALL} & \textbf{Avg.} \\ \midrule
\multirow{4}{*}{\textbf{mDeBERTa-v3}} & 
\textbf{CQ\#E} & 32.53 & 39.75 & 36.14 & 36.14 & 36.14   \\ 
&\textbf{CQP\#E} & 27.58 & 35.63 & 31.03 & 37.93 & 33.04  \\ 
&\textbf{Q\#CE} & 28.73  & 27.58 & 29.88 & 35.63  & 30.45        \\ 
&\textbf{QP\#CE} & 25.28 & 32.18 & 33.33 & 34.48 & 31.31       \\ \midrule
\multirow{4}{*}{\textbf{XLM-R-base}} & 
\textbf{CQ\#E} & 25.30                  & 42.16                         & 28.91                         & 33.73                         & 32.52        \\ 
& \textbf{CQP\#E} & 26.43 & 43.67 & 33.33                         & 34.48                         & 34.47       \\
& \textbf{Q\#CE} & 27.58 & 32.18                         & 29.88                         & 35.63                         & 31.31       \\ 
& \textbf{QP\#CE} & 28.73                         & 33.33 & 29.88                         & 32.18 & 31.03         \\ \midrule
\multirow{4}{*}{\textbf{XLM-R-large}}   & 
\textbf{CQ\#E} & 37.34 & 40.96                         & 40.96                         & 40.96 & 40.05        \\ 
& \textbf{CQP\#E} & 32.18                         & 39.08                         & 41.37 & 37.93 & 37.64         \\ 
& \textbf{Q\#CE} & 27.58                         & 32.18                         & 29.88                         & 35.63                         & 31.31       \\ 
& \textbf{QP\#CE} & 26.43                         & 35.63
				& 39.08                         & 33.33
				& 33.61       \\ \midrule
\multirow{4}{*}{\textbf{XLM-R-GALEN}}   & 
\textbf{CQ\#E} & 27.71                         & 30.12
			   & 26.50                          & 31.32                         & 28.91       \\ 
& \textbf{CQP\#E} & 25.28                         & 27.58                         & 22.98                         & 28.73                         & 26.14       \\ 
& \textbf{Q\#CE} & 18.39                         & 17.24                         & 21.83                         & 26.43                         & 20.97       \\ 
& \textbf{QP\#CE} & 19.54                         & 22.98
				& 24.13                         & 21.83
				& 22.12         \\ \midrule
\multirow{4}{*}{\textbf{eriBERTa} }     & 
\textbf{CQ\#E} & 34.93                         & 34.93                         & 37.34                         & 33.73                         & 35.23       \\ 
& \textbf{CQP\#E} & 27.58                         & 29.88                         & 28.73                         & 32.18                         & 29.59       \\ 
& \textbf{Q\#CE} & 29.88                         & 39.08                         & 28.73                         & 41.37                         & 34.76        \\
& \textbf{QP\#CE} & 33.33                         & 33.33
				& 28.73                         & 33.33
				& 32.18         \\ \bottomrule
\end{tabular}
}
\caption{Exact match results of the multilingual models on different \emph{CasiMedicos} dataset variants where \textbf{CM} = Fine-tuning on \emph{CasiMedicos} dataset, \textbf{SQAC} = Fine-tuning on SQAC dataset, \textbf{SQES} = Fine-tuning on SQuAD-es dataset. \textbf{C}: Clinical Case; \textbf{Q}: Question; \textbf{E}: Medical Doctor's Explanations; \textbf{P}:
Possible Answers. \textbf{Fragment before \#}: generated question;
\textbf{after \#}: generated context from which to extract the correct
explanation.}
\label{table:casimedicosmultiexact}
\end{table}

\begin{table}[h]
\centering
\footnotesize
\resizebox{12cm}{!} {
\begin{tabular}{c|l|c|c|c|c|c}
\toprule
 \multicolumn{2}{l}{} & \textbf{CM} & \textbf{SQAC+CM} & \textbf{SQES+CM} & \textbf{ALL} & \textbf{Avg.} \\ \midrule
\multirow{4}{*}{\textbf{IXAmBERT}} & 
\textbf{CQ\#E}      & 64.47       & 64.89            & 67.40               & 66.11                    & 65.71       \\
& \textbf{CQP\#E} & 64.37       & 67.28            & 63.42               & 67.81                    & 65.72         \\
& \textbf{Q\#CE}      & 64.69       & 68.49            & 61.91               & 64.13                    & 64.80        \\
& \textbf{QP\#CE} & 64.88       & 67.69            & 66.57               & 65.39
				& 66.13       \\ \midrule

\multirow{4}{*}{\textbf{MarIA-base}}   &
\textbf{CQ\#E}      & 53.51       & 66.11            & 70.25               & 67.47                    & 64.33        \\ 
& \textbf{CQP\#E} & 53.04       & 67.37            & 73.82               & 65.94                    & 65.04       \\ 
& \textbf{Q\#CE}      & 40.65       & 63.08            & 67.24               & 66.51                    & 59.37         \\ 
& \textbf{QP\#CE} & 38.80        & 59.51            & 66.68               & 66.82 & 57.95       \\ \midrule
\multirow{4}{*}{\textbf{MarIA-large}}    & 
\textbf{CQ\#E}      & 59.18       & 68.75            & 67.69               & 67.21                    & 65.70       \\
& \textbf{CQP\#E} & 58.64       & 68.08            & 64.16               & 70.46                    & 65.33        \\
& \textbf{Q\#CE}      & 50.40        & 67.74            & 69.24               & 63.55                    & 62.73       \\
& \textbf{QP\#CE} & 52.59       & 63.13            & 66.71               & 63.40
				& 61.45       \\ \midrule
\multirow{4}{*}{\textbf{beto} }      & 
\textbf{CQ\#E}      & 67.34       & 66.08            & 69.27               & 68.58                    & 67.81       \\
& \textbf{CQP\#E} & 65.71       & 67.22            & 69.75               & 69.33 & 68.00       \\
& \textbf{Q\#CE}      & 61.36       & 60.71            & 68.85               & 67.26                    & 64.54        \\
& \textbf{QP\#CE} & 62.92       & 62.55            & 67.36               & 66.13
				& 64.74         \\ \midrule
 
\multirow{4}{*}{\textbf{IXAes}}  & 
\textbf{CQ\#E}      & 68.18       & 67.15            & 65.90                & 69.74                    & 67.74       \\
& \textbf{CQP\#E} & 66.95       & 67.74            & 68.50               & 64.96                    & 67.03       \\
& \textbf{Q\#CE}      & 62.34       & 67.93            & 64.53               & 60.43                    & 63.80       \\
& \textbf{QP\#CE} & 60.81       & 64.93            & 60.35               & 62.85 & 62.23        \\ \midrule

\multirow{4}{*}{\textbf{BSC-BIO-EHR}}       & 
\textbf{CQ\#E}      & 61.99       & 66.18            & 72.05               & 69.64                    & 67.46        \\
& \textbf{CQP\#E} & 58.27       & 67.15            & 69.43               & 70.28                    & 66.28       \\ 
& \textbf{Q\#CE}      & 60.55       & 66.11            & 65.42               & 66.29                    & 64.59       \\
& \textbf{QP\#CE} & 61.69       & 63.47            & 66.04               & 68.93
				& 65.03       \\ \midrule
\multirow{4}{*}{\textbf{BSC-BIO}}    & 
\textbf{CQ\#E}      & 63.87       & 66.66            & 69.84               &
72.30                     & 68.16       \\
& \textbf{CQP\#E} & 64.11       & 66.91            & 66.28               & 71.46                    & 67.19         \\
& \textbf{Q\#CE}      & 65.44       & 66.53            & 69.79               & 68.66                    & 67.60        \\
& \textbf{QP\#CE} & 61.22       & 63.37            & 69.82               & 63.99
				& 64.60          \\ \bottomrule
\end{tabular}
}
\caption{F1-score results of Spanish monolingual models on different \emph{CasiMedicos} dataset variants where \textbf{CM} = Fine-tuning on \emph{CasiMedicos} dataset, \textbf{SQAC} = Fine-tuning on SQAC dataset, \textbf{SQES} = Fine-tuning on SQuAD-es dataset. \textbf{C}: Clinical Case; \textbf{Q}: Question; \textbf{E}: Medical Doctor's Explanations; \textbf{P}:
Possible Answers. \textbf{Fragment before \#}: generated question;
\textbf{after \#}: generated context from which to extract the correct
explanation.}
\label{table:casimedicosmono}
\end{table}

\begin{table}[h]
\centering
\footnotesize
\resizebox{12cm}{!} {
\begin{tabular}{c|l|c|c|c|c|c}
\toprule
 \multicolumn{2}{l}{} & \textbf{CM} & \textbf{SQAC+CM} & \textbf{SQES+CM} & \textbf{ALL} & \textbf{Avg.} \\ \midrule
\multirow{4}{*}{\textbf{IXAmBERT}}  & 
\textbf{CQ\#E}                & 27.71       & 30.12                         & 34.93               & 33.73                         & 31.62       \\ 
& \textbf{CQP\#E}                & 33.33       & 28.73                         & 27.58               & 36.78                         & 31.60        \\ 
& \textbf{Q\#CE}                & 26.43       & 33.33                         & 22.98               & 25.28                         & 27.00        \\ 
& \textbf{QP\#CE}                & 27.56       & 32.18                         & 29.88               & 31.03                         & 30.16       \\ \midrule

\multirow{4}{*}{\textbf{MarIA-base}}   & 
\textbf{CQ\#E}                & 26.50        & 26.50                          & 37.34               & 34.93 & 31.31       \\ 
& \textbf{CQP\#E}                & 22.98       & 27.58 & 40.22               & 29.88                         & 30.16        \\ 
& \textbf{Q\#CE}                & 11.49       & 25.28                         & 33.33               & 33.33                         & 25.85       \\ 
& \textbf{QP\#CE}                & 14.94       & 19.54                         & 28.73               & 32.18                         & 23.84       \\ \midrule

\multirow{4}{*}{\textbf{MarIA-large}} & 
\textbf{CQ\#E}                & 24.09       & 32.53                         & 32.53               & 33.73                         & 30.72         \\ 
 
& \textbf{CQP\#E}                & 22.98       & 37.93                         & 31.03               & 33.33                         & 31.31       \\ 
 
& \textbf{Q\#CE}                & 14.94       & 32.18                         & 32.18               & 33.33                         & 28.15       \\ 
 
& \textbf{QP\#CE}                & 20.68       & 31.03                         & 34.48               & 28.73                         & 28.73         \\ \midrule
 
\multirow{4}{*}{\textbf{beto}}  & 
\textbf{CQ\#E}                & 30.12       & 27.71                         & 36.14               & 32.53                         & 31.62        \\ 
& \textbf{CQP\#E}                & 26.43       & 31.03                         & 35.63               & 34.48                         & 31.89       \\ 
& \textbf{Q\#CE}                & 22.98       & 25.28                         & 32.18               & 32.18                         & 28.15        \\ 
& \textbf{QP\#CE}                & 22.98       & 25.28                         & 34.48               & 27.58                         & 27.58         \\ \midrule
 
\multirow{4}{*}{\textbf{IXAes}}      &
\textbf{CQ\#E}                & 28.91       & 31.32                         & 28.91               & 33.73                         & 30.71       \\ 
 
& \textbf{CQP\#E}                & 31.03       & 31.03                         & 28.73               & 25.28                         & 29.01       \\ 
 
& \textbf{Q\#CE}                & 27.58       & 27.58                         & 27.58               & 26.43                         & 27.29       \\ 
 
& \textbf{QP\#CE}                & 28.73       & 29.88                         & 24.13               & 26.43                         & 27.29       \\ \midrule
 
\multirow{4}{*}{\textbf{BSC-BIO-EHR}}  & 
\textbf{CQ\#E}                & 33.73       & 30.12                         & 40.49               & 32.53                         & 34.21       \\ 
 
& \textbf{CQP\#E}                & 28.73       & 32.18                         & 39.08               & 32.18                         & 33.04       \\
 
& \textbf{Q\#CE}                & 29.88       & 32.18                         & 31.03               & 29.88                         & 31.03         \\
 
& \textbf{QP\#CE}                & 26.43       & 28.73                         & 25.28               & 31.03                         & 27.86      \\ \midrule
 
\multirow{4}{*}{\textbf{BSC-BIO}}  & 
\textbf{CQ\#E}                & 31.32       & 31.32                         & 33.73               & 38.55                         & 33.73         \\ 
 
& \textbf{CQP\#E}                & 33.33       & 35.63                         & 36.78               & 37.93                         & 35.91       \\ 
 
& \textbf{Q\#CE}                & 31.03       & 31.03                         & 34.48               & 35.63                         & 33.04       \\ 
 
& \textbf{QP\#CE}                & 27.58       & 29.88                         & 32.18               & 31.03                         & 30.16       \\ \bottomrule

\end{tabular}
}
\caption{Exact match results of Spanish monolingual models on different \emph{CasiMedicos} dataset variants where \textbf{CM} = Fine-tuning on \emph{CasiMedicos} dataset, \textbf{SQAC} = Fine-tuning on SQAC dataset, \textbf{SQES} = Fine-tuning on SQuAD-es dataset. \textbf{C}: Clinical Case; \textbf{Q}: Question; \textbf{E}: Medical Doctor's Explanations; \textbf{P}:
Possible Answers. \textbf{Fragment before \#}: generated question; \textbf{after \#}: generated context from which to extract the correct explanation.}
\label{table:casimedicosmonoexact}
\end{table}

\begin{table}[h]
\centering
\footnotesize
\resizebox{12cm}{!} {
\begin{tabular}{c|l|c|c|c|c|c}
\toprule
 \multicolumn{2}{l}{} & \textbf{MA} & \textbf{SQAC+MA} & \textbf{SQES+MA} & \textbf{ALL} & \textbf{Avg.} \\ \midrule
\multirow{4}{*}{\textbf{mDeBERTa-v3}}       & 
\textbf{CQ\#E}      & 79.36       & 82.57            & 79.40                & 82.23                    & 80.89         \\ 
& \textbf{CQP\#E} & 77.82       & 81.51            & 77.32               & 80.80                     & 79.36       \\ 
& \textbf{Q\#CE}      & 78.26       & 79.51            & 79.99               & 79.87                    & 79.40       \\ 
& \textbf{QP\#CE} & 71.49       & 72.19            & 74.39               & 73.72 & 72.94       \\ \midrule
 
\multirow{4}{*}{\textbf{XLM-R-base}}   & 
\textbf{CQ\#E}      & 78.39       & 83.52            & 81.98               & 83.83                    & 81.93         \\ 
& \textbf{CQP\#E} & 82.23       & 81.89            & 82.90                & 77.84                    & 81.21        \\ 
& \textbf{Q\#CE}      & 81.03       & 80.66            & 82.89               & 80.07                    & 81.16       \\ 
& \textbf{QP\#CE} & 74.25       & 76.39            & 77.73               & 77.95 & 76.58         \\ \midrule
 
\multirow{4}{*}{\textbf{XLM-R-large}}   & 
\textbf{CQ\#E}      & 82.20        & 84.30             & 85.92               & 82.27                    & 83.67       \\ 
& \textbf{CQP\#E} & 81.25       & 83.63            & 79.68               & 82.78                    & 81.83        \\ 
& \textbf{Q\#CE}      & 81.03       & 80.66            & 82.89               & 80.07                    & 81.16       \\ 
& \textbf{QP\#CE} & 74.49       & 77.15            & 77.46               & 79.36 & 77.11        \\ \midrule
 
\multirow{4}{*}{\textbf{XLM-R-GALEN}}       & 
\textbf{CQ\#E}      & 74.08       & 72.72            & 79.85               & 79.89                    & 76.63       \\ 
& \textbf{CQP\#E} & 71.72       & 76.47            & 78.42               & 82.02                    & 77.15       \\ 
& \textbf{Q\#CE}      & 66.88       & 67.55            & 73.42               & 74.84                    & 70.67       \\
& \textbf{QP\#CE} & 70.19       & 68.92            & 74.91               & 71.50 & 71.38         \\ \midrule

\multirow{4}{*}{\textbf{eriBERTa}}        &
\textbf{CQ\#E}      & 84.92       & 85.89            & 85.25               & 84.50                     & 85.14         \\ 
&\textbf{CQP\#E} & 82.66       & 85.45            & 84.33               & 85.42                    & 84.46        \\ 
&\textbf{Q\#CE}      & 82.08       & 83.70             & 84.14               & 87.03                    & 84.23       \\ 
& \textbf{QP\#CE} & 80.27       & 82.56            & 84.29               & 83.08
				& 82.55         \\ \bottomrule
\end{tabular}
}
\caption{F1-score results of the multilingual models on different \emph{MIR Asturias} dataset variants where \textbf{MA} = Fine-tuning on \emph{MIR Asturias} dataset, \textbf{SQAC} = Fine-tuning on SQAC dataset, \textbf{SQES} = Fine-tuning on SQuAD-es dataset. \textbf{C}: Clinical Case; \textbf{Q}: Question; \textbf{E}: Medical Doctor's Explanations; \textbf{P}:
Possible Answers. \textbf{Fragment before \#}: generated question; \textbf{after \#}: generated context from which to extract the correct explanation.}
\label{table:mirmulti}
\end{table}

\begin{table}[h]
\centering
\footnotesize
\resizebox{12cm}{!} {
\begin{tabular}{c|l|c|c|c|c|c}
\toprule
 \multicolumn{2}{l}{} & \textbf{MA} & \textbf{SQAC+MA} & \textbf{SQES+MA} & \textbf{ALL} & \textbf{Avg.} \\ \midrule
\multirow{4}{*}{\textbf{mDeBERTa-v3}}   & 
\textbf{CQ\#E} & 67.52                         & 70.08                         & 65.81                         & 68.37                         & 67.94        \\ 
& \textbf{CQP\#E} & 65.81                         & 67.52                         & 65.81                         & 65.81                         & 66.23       \\ 
& \textbf{Q\#CE} & 63.24                         & 66.66                         & 68.37                         & 66.66                         & 66.23      \\ 
& \textbf{QP\#CE} & 59.82                         & 60.68
				& 59.82                         & 58.11
				& 59.60       \\ \midrule
\multirow{4}{*}{\textbf{XLM-R-base}}   & 
\textbf{CQ\#E} & 62.39                         & 67.52                         & 65.81                         & \cellcolor[HTML]{FFFFFF}70.08 & 66.45         \\ 
& \textbf{CQP\#E} & 64.95                         & \cellcolor[HTML]{FFFFFF}68.37 & 67.52                         & 63.24                         & 66.02         \\
& \textbf{Q\#CE} & 64.10                          & 66.66                         & 67.52                         & 65.81                         & 66.02       \\ 
& \textbf{QP\#CE} & 59.82                         & 63.24 & 59.82                         & 61.53 & 61.10       \\ \midrule
 
\multirow{4}{*}{\textbf{XLM-R-large}}   & 
\textbf{CQ\#E} & \cellcolor[HTML]{FFFFFF}68.37 & 70.94                         & 71.79                         & \cellcolor[HTML]{FFFFFF}69.23 & 70.08       \\
& \textbf{CQP\#E} & 65.81                         & 70.94                         & \cellcolor[HTML]{FFFFFF}65.81 & \cellcolor[HTML]{FFFFFF}68.37 & 67.73       \\
& \textbf{Q\#CE} & 64.10                          & 66.66                         & 67.52                         & 65.81                         & 66.02      \\
& \textbf{QP\#CE} & 59.82                         & 63.24 & 62.39                         & 62.39 & 61.96         \\ \midrule
 
\multirow{4}{*}{\textbf{XLM-R-GALEN}}      & 
\textbf{CQ\#E} & 52.13                         & 53.84                         & 60.68                         & 59.82                         & 56.61       \\ 
& \textbf{CQP\#E} & 47.86                         & 54.70                          & 58.11                         & 63.24                         & 55.97       \\
& \textbf{Q\#CE} & 37.60                          & 45.29                         & 52.99                         & 53.84                         & 47.43         \\
& \textbf{QP\#CE} & 41.02                         & 43.58 & 54.70                          & 50.42 & 47.43         \\ \midrule
 
\multirow{4}{*}{\textbf{eriBERTa}}     & 
\textbf{CQ\#E} & 69.23                         & 71.79                         & 71.79                         & 66.66                         & 69.86       \\

& \textbf{CQP\#E} & 69.23                         & 68.37                         & 69.23                         & 70.94                         & 69.44       \\

& \textbf{Q\#CE} & 64.95                         & 63.24                         & 67.52                         & 72.64                         & 67.08       \\

& \textbf{QP\#CE} & 64.95                         & 67.52 & 69.23 & 65.81 & 66.87       \\ \bottomrule
\end{tabular}
}
\caption{Exact match results of the multilingual models on different \emph{MIR Asturias} dataset variants where \textbf{MA} = Fine-tuning on \emph{MIR Asturias} dataset, \textbf{SQAC} = Fine-tuning on SQAC dataset, \textbf{SQES} = Fine-tuning on SQuAD-es dataset. \textbf{C}: Clinical Case; \textbf{Q}: Question; \textbf{E}: Medical Doctor's Explanations; \textbf{P}:
Possible Answers. \textbf{Fragment before \#}: generated question; \textbf{after \#}: generated context from which to extract the correct explanation.}
\label{table:mirmultiexact}
\end{table}

\begin{table}[h]
\centering
\footnotesize
\resizebox{12cm}{!} {
\begin{tabular}{c|l|c|c|c|c|c}
\toprule
 \multicolumn{2}{l}{} & \textbf{MA} & \textbf{SQAC+MA} & \textbf{SQES+MA} & \textbf{ALL} & \textbf{Avg.} \\ \midrule
\multirow{4}{*}{\textbf{IXAmBERT}}          & 
\textbf{CQ\#E}      & 82.66       & 86.14            & 83.86               & 83.54                    & 84.05         \\ 
& \textbf{CQP\#E} & 82.41       & 86.14            & 82.73               & 83.84                    & 83.78         \\ 
& \textbf{Q\#CE}      & 78.83       & 82.32            & 79.30                & 80.84                    & 80.32       \\ 
& \textbf{QP\#CE} & 77.56       & 80.18            & 78.89               & 75.03 & 77.91        \\ \midrule
 
\multirow{4}{*}{\textbf{MarIA-base}}   & 
\textbf{CQ\#E}      & 70.85       & 82.53            & 82.71               & 82.01                    & 79.52        \\ 
& \textbf{CQP\#E} & 75.52       & 83.79            & 81.97               & 81.47                    & 80.68       \\ 
& \textbf{Q\#CE}      & 47.80        & 79.05            & 80.94               & 76.18                    & 70.99       \\ 
& \textbf{QP\#CE} & 52.95       & 76.09            & 75.87               & 74.93 & 69.96         \\ \midrule
 
\multirow{4}{*}{\textbf{MarIA-large}}  & 
\textbf{CQ\#E}      & 79.98       & 83.66            & 84.80                & 82.26                    & 82.67        \\ 
& \textbf{CQP\#E} & 80.54       & 82.90             & 83.94               & 80.73                    & 82.02       \\ 
& \textbf{Q\#CE}      & 76.26       & 80.94            & 75.18               & 81.13                    & 78.37       \\ 
&\textbf{QP\#CE} & 72.55       & 75.77            & 77.59               & 77.11 & 75.75        \\ \midrule
 
\multirow{4}{*}{\textbf{beto}}          & 
\textbf{CQ\#E}      & 82.00          & 82.69            & 84.92               & 82.24                    & 82.96       \\ 
& \textbf{CQP\#E} & 82.03       & 83.39            & 83.32               & 83.25                    & 82.99       \\ 
& \textbf{Q\#CE}      & 81.89       & 82.18            & 79.15               & 78.83                    & 80.51       \\ 
& \textbf{QP\#CE} & 76.73       & 80.55            & 79.93               & 80.24 & 79.36       \\ \midrule
 
\multirow{4}{*}{\textbf{IXAes}}   & 
\textbf{CQ\#E}      & 82.37       & 86.56            & 86.34               & 84.69                    & 84.99         \\
& \textbf{CQP\#E} & 79.98       & 86.85            & 84.39               & 84.80 & 84.00       \\ 
& \textbf{Q\#CE}      & 81.69       & 83.34            & 81.58               & 79.11                    & 81.43         \\ 
& \textbf{QP\#CE} & 79.14       & 78.02            & 81.15               & 78.95 & 79.31        \\ \midrule
 
\multirow{4}{*}{\textbf{BSC-BIO-EHR}}          &
\textbf{CQ\#E}      & 79.47       & 83.82            & 83.17               & 81.61                    & 82.01       \\ 
& \textbf{CQP\#E} & 75.48       & 84.58            & 84.73               & 85.87                    & 82.66        \\ 
& \textbf{Q\#CE}      & 73.99       & 81.17            & 79.18               & 84.37                    & 79.67      \\ 
&\textbf{QP\#CE} & 74.89       & 80.62            & 80.80                & 82.13 & 79.61         \\ \midrule
 
\multirow{4}{*}{\textbf{BSC-BIO}}      &
\textbf{CQ\#E}      & 82.74       & 81.56            & 83.56               & 83.47                    & 82.83       \\ 
& \textbf{CQP\#E} & 82.61       & 83.68            & 82.61               & 84.31                    & 83.30       \\ 
& \textbf{Q\#CE}      & 79.92       & 82.78            & 82.40                & 82.42                    & 81.88         \\ 
& \textbf{QP\#CE} & 78.57       & 81.61            & 82.46               & 83.26
				& 81.47        \\ \bottomrule

\end{tabular}
}
\caption{F1-score results of Spanish monolingual models on different \emph{MIR Asturias} dataset variants where \textbf{MA} = Fine-tuning on \emph{MIR Asturias} dataset, \textbf{SQAC} = Fine-tuning on SQAC dataset, \textbf{SQES} = Fine-tuning on SQuAD-es dataset. \textbf{C}: Clinical Case; \textbf{Q}: Question; \textbf{E}: Medical Doctor's Explanations; \textbf{P}:
Possible Answers. \textbf{Fragment before \#}: generated question; \textbf{after \#}: generated context from which to extract the correct explanation.}
\label{table:mirmono}
\end{table}

\begin{table}[h]
\centering
\footnotesize
\resizebox{12cm}{!} {
\begin{tabular}{c|l|c|c|c|c|c}
\toprule
 \multicolumn{2}{l}{} & \textbf{MA} & \textbf{SQAC+MA} & \textbf{SQES+MA} & \textbf{ALL} & \textbf{Avg.} \\ \midrule
\multirow{4}{*}{\textbf{IXAmBERT}}       & 
\textbf{CQ\#E} & 61.53       & 70.08                         & 64.95               & 62.39                         & 64.73                                                    \\ 
& \textbf{CQP\#E} & 65.81       & 67.52                         & 64.95               & 67.52                         & 66.45                                                      \\ 
& \textbf{Q\#CE} & 58.11       & 61.53                         & 58.11               & 58.11                         & 58.96                                                     \\ 
& \textbf{QP\#CE} & 60.68       & 59.82                         & 60.68 & 50.42                        & 57.90 \\ \midrule
 
\multirow{4}{*}{\textbf{MarIA-base}}     & 
\textbf{CQ\#E} & 55.55       & 63.24                         & 64.10                & 65.81 & 62.17                                                     \\ 
& \textbf{CQP\#E} & 58.97       & 64.95 & 61.53               & 62.39                         & 61.96                                                      \\ 
& \textbf{Q\#CE} & 31.62       & 55.55                         & 62.39               & 53.84                         & 50.85                                                      \\ 
& \textbf{QP\#CE} & 36.75       & 56.41                         & 54.70
				& 53.84                         & 50.42
				\\ \midrule
 
\multirow{4}{*}{\textbf{MarIA-large}}   & 
\textbf{CQ\#E} & 61.53       & 70.08                         & 72.64               & 68.37                         & 68.15                                                     \\ 
 
& \textbf{CQP\#E} & 60.68       & 65.81                         & 65.81               & 66.66                         & 64.74                                                      \\ 
 
& \textbf{Q\#CE} & 46.15       & 58.11                         & 54.70                & 61.53                         & 55.12                                                    \\
 
& \textbf{QP\#CE} & 43.58       & 58.11                         & 60.68
				& 63.24                         & 56.40
				\\ \midrule
 
\multirow{4}{*}{\textbf{beto}}       & 
\textbf{CQ\#E} & 66.66       & 64.95                         & 68.37               & 65.81                         & 66.44                                                    \\ 
& \textbf{CQP\#E} & 64.10        & 62.39                         & 65.81               & 67.52                         & 64.95                                                     \\ 
& \textbf{Q\#CE} & 64.95       & 64.95                         & 63.24               & 58.97                         & 63.02                                                    \\
& \textbf{QP\#CE} & 59.82       & 64.95                         & 63.24 & 58.97                         & 61.74 \\ \midrule
 
\multirow{4}{*}{\textbf{IXAes}}    &
\textbf{CQ\#E} & 66.66       & 71.79                         & 70.94               & 67.52                         & 69.22                                                    \\
 
& \textbf{CQP\#E} & 58.97       & 69.23                         & 69.23               & 67.52                         & 66.23                                                   \\
 
& \textbf{Q\#CE} & 64.95       & 67.52                         & 60.68               & 60.68                         & 63.45                                                    \\
 
& \textbf{QP\#CE} & 60.68       & 58.97                         & 63.24 & 62.39                         & 61.32 \\ \midrule
 
\multirow{4}{*}{\textbf{BSC-BIO-EHR}}    & 
\textbf{CQ\#E} & 61.53       & 67.52                         & 69.23               & 60.68                         & 64.74                                                      \\ 
 
& \textbf{CQP\#E} & 59.82       & 66.66                         & 70.08               & 67.52                         & 66.02                                                      \\ 
 
& \textbf{Q\#CE} & 56.41       & 62.39                         & 63.24               & 62.39                         & 61.10                                                    \\ 
 
& \textbf{QP\#CE} & 58.11       & 61.53                         & 62.39
				& 60.68                         & 60.67
				\\ \midrule
 
\multirow{4}{*}{\textbf{BSC-BIO}}    &
\textbf{CQ\#E} & 64.95       & 64.10                          & 64.95               & 66.66                         & 65.16                                                     \\ 
 
& \textbf{CQP\#E} & 68.37       & 64.95                         & 66.66               & 67.52                         & 66.87                                                     \\
 
& \textbf{Q\#CE} & 64.10        & 64.95                         & 64.95               & 63.24                         & 64.31                                                      \\ 
 
& \textbf{QP\#CE} & 59.82       & 62.39                         & 60.68
				& 61.53                         & 61.10
				\\ \bottomrule
\end{tabular}
}
\caption{Exact match results of Spanish monolingual models on different \emph{MIR Asturias} dataset variants where \textbf{MA} = Fine-tuning on \emph{MIR Asturias} dataset, \textbf{SQAC} = Fine-tuning on SQAC dataset, \textbf{SQES} = Fine-tuning on SQuAD-es dataset. \textbf{C}: Clinical Case; \textbf{Q}: Question; \textbf{E}: Medical Doctor's Explanations; \textbf{P}:
Possible Answers. \textbf{Fragment before \#}: generated question; \textbf{after \#}: generated context from which to extract the correct explanation.}
\label{table:mirmonoexact}
\end{table}

\end{document}